\documentclass[12pt]{article}
\usepackage[english]{babel}
\usepackage{adjustbox}

\usepackage[letterpaper,top=2cm,bottom=2cm,left=2cm,right=2cm,marginparwidth=1.75cm]{geometry}

\usepackage{algorithm}
\usepackage{algpseudocode}
\usepackage{tabularx}
\usepackage{multirow}
\usepackage{booktabs}
\usepackage{tabularx}
\usepackage{adjustbox}
\usepackage{xcolor}
\usepackage{threeparttable}
\usepackage{lscape}
\usepackage{appendix}
\usepackage{lineno}
\usepackage{subfig}

\usepackage{amsmath}
\usepackage{amssymb}
\usepackage[colorlinks=true, allcolors=blue]{hyperref}
\usepackage{array,calc}
\usepackage{lscape}
\usepackage{float}
\usepackage[nameinlink,capitalize]{cleveref}
\newcommand\independent{\protect\mathpalette{\protect\independenT}{\perp}}\def\independenT#1#2{\mathrel{\rlap{$#1#2$}\mkern2mu{#1#2}}}
\usepackage{soul}

\newtheorem{definition}{Definition}

\usepackage{booktabs} % for toprule, midrule, bottomrule in tables
\usepackage{authblk} % for multiple authors and affiliations
\usepackage{graphicx} % to inlcude graphics with \includegraphics
\usepackage[
    backend=biber,
    style=apa,
  ]{biblatex}
\usepackage{caption}
\author[1]{Kimia Kamal\thanks{kimia.kamal@torontomu.ca}}

\author[1]{Bilal Farooq\thanks{bilal.farooq@torontomu.ca (corresponding author)}}

\affil[1]{Laboratory of Innovations in Transportation (LiTrans),

Toronto Metropolitan University, Toronto, Canada}

\title{\textbf{A deep causal inference model for interpretable travel behaviour analysis}\\Published in: Transportation Research Part C: Emerging Technologies}
\date{}

\begin{document}
\maketitle

\begin{abstract}

Transportation policy assessment often involves causal questions, yet traditional travel behaviour lacks the ability of distinguishing causal association from non-causal correlations. We present the \emph{deep CAusal infeRence mOdel for traveL behavIour aNAlysis (CAROLINA)}, %This novel framework integrates causality analysis with traditional discrete choice models and leverages deep learning algorithms. 
a novel framework that addresses both observed and unobserved confounders in discrete choice modelling through causal discovery and the strengths of deep neural networks. Within this interpretable framework, we also introduce a Generative Counterfactual model for forecasting individuals' behaviour by adapting the Normalizing Flow method. Through the case studies of virtual reality-based pedestrian crossing behaviour, revealed preference travel behaviour from London, and synthetic data, we demonstrate the effectiveness of our proposed models in isolating causal relationships from confounding associations. Our proposed causal framework enhances predictive accuracy by employing data-driven algorithms, while maintaining interpretability. In addition, the counterfactual results show that intervention mechanisms that can reduce pedestrian stress levels lead to a 35.35\% increase in individuals experiencing shorter waiting times. External interventions that reduce travel distances of non-mandatory trips in London lead to a 47\% increase in the possible use of active and public transit modes.
\end{abstract}


\section{Introduction}

%Human behaviour is a complex phenomenon influenced by a variety of factors, conditions, situations, and environments. In transportation, the analysis of travel behaviour plays an essential role in supporting modellers and decision-makers to effectively plan, design, and operate transportation infrastructure and associated services. 
In recent years, the emergence of big data, deep learning, and data science have opened new possibilities for travel behaviour analysis. The considerable ability of these flexible models to capture underlying real-world behaviour from the data has resulted in more accurate travel behaviour prediction models \parencite{hillel2021systematic, wang2018machine, kalatian2021decoding}. Despite the noticeable success in the development of advanced travel behaviour models, there are two important concerns about these frameworks \parencite{karlaftis2011statistical, brathwaite2018causal}: 1) lack of interpretability, known as black-box nature, and 2) lack of explicit incorporation and quantification of causality---correlation does not mean causation. 
Various studies have focused on overcoming the first aforementioned restriction of deep learning algorithms in the context of travel behavioural modelling by applying eXplainable Artificial Intelligence (XAI) methods or integrating deep learning algorithms into traditional behavioural models \parencite{wong2021reslogit, kamal2024ordinal, wang2020deep, sifringer2020enhancing, kalatian2021decoding}\footnote{{A more detailed discussion on these approaches can be found in review papers, such as \cite{kalatian2024travel, van2022choice}.}}. These studies have tried to leverage the strengths of both modelling techniques, achieving a balanced integration without sacrificing the theoretical underpinning. However, in terms of causality, these models are still criticized for not accounting for the effect of variables that confound the causal impacts \parencite{brathwaite2018causal}. In human behaviour analysis, individual characteristics and their preference towards specific behaviour exemplify prominent instances of observed and unobserved confounding variables. These confounding variables can interact with both explanatory variables and an outcome, leading to both causal and non-causal associations between variables \parencite{pearl2009causal}. Generally, discrete choice models (DCMs) and deep learning algorithms measure the combination of causal and non-causal association without specifically isolating them individually \parencite{brathwaite2018causal, athey2019generalized, kamal2022debiased}. %{In the context of forecasting, the majority of counterfactual models in transportation rely on the Stable Unit Treatment Value Assumption (SUTVA)} \parencite{brathwaite2018causal}{. This means that the treatment applied to one individual does not affect the outcome of another individual. Thus, under SUTVA, policies are evaluated on individuals in isolation. However, this assumption is not always true, e.g., providing free parking to a subset of individuals can have a ``spillover'' effect on others. 
When evaluating policies, existing approaches heavily rely on quasi-experiments such as Difference-in-Differences (DiD) or Regression Discontinuity Design (RDD). However, these methods suffer from over-reliance on the parallel trends assumption, are incapable of properly accounting for dynamic and staggered introduction of policies, identify the local average treatment effect, and underestimate standard error due to serial correlations \parencite{rokicki2018inference, egerod2024many, o2016approaches}.

In order to bridge these gaps and provide a better forecasting tool for the analysis of urban plans and policies, we aim to develop a novel framework that simultaneously employs traditional DCMs, deep learning algorithms, and causal inference techniques. Briefly, our research investigates two key questions:

\begin{itemize}
    \item What are ways to mitigate the influence of observed and unobserved confounding variables?
    \item Can the forecasting be enhanced by incorporating causality-based methods?
\end{itemize}

To address these questions, we propose the \emph{deep CAusal infeRence mOdelling for traveL behavIour aNAlysis (CAROLINA)} framework. First, a probabilistic behaviour modelling structure within the \emph{CAROLINA} framework addresses the observed and unobserved confounders by leveraging causal learning techniques and interpretable deep learning models \parencite{wong2021reslogit, kamal2024ordinal}. 
%Our approach to employing deep neural networks aligns with our previous efforts that ensure model interpretability \parencite{wong2021reslogit, kamal2024ordinal}. 
%It is worth noting that \emph{CAROLINA} is still classified as a probabilistic behaviour model, where our focus is on causal association after accounting for confounders, rather than a combination of causal and non-causal association. 
Secondly, in order to provide insights into the effects of manipulation on the system, in the forecasting step, we incorporate the fact that observational and post-intervention systems are not identically distributed. In fact, once an external intervention is imposed on the system, it induces changes in the behaviour pattern and causal structure of the model. This results in the emergence of interventional or counterfactual distributions. Furthermore, counterfactual analysis offers valuable insight at the individual level by considering how people imagine themselves in a counterfactual world, ask counterfactual-based questions, and make optimal decisions. For instance, pedestrians may unconsciously ask themselves ``what would happen if I crossed the road immediately without attention to vehicles'' or employees may think daily, ``imagine if we would get used to using active modes! We propose a novel approach for estimating counterfactual human behaviour by learning latent constructs and considering intervention impact based on causality analysis. Three different datasets, including stated preference \parencite{farooq2018virtual}, revealed preference \parencite{hillel2018recreating}, and simulated data, are used to develop detailed analysis.
%In short, the \emph{CAROLINA} framework aims to advance causal analysis in travel behaviour modelling by explicitly accounting for observed and unobserved confounders, enhancing predictive accuracy and improving forecasting analysis. 
Accordingly, the work contributes to the transportation literature in the following ways:

\begin{itemize}

    \item Formulation of an interpretable deep causality-based behavioural model;
    \item Formulation of a generative counterfactual model for forecasting, which, to the best of our knowledge, represents the first such investigation in transportation research. 
    
\end{itemize}

%In this study, we use three datasets to show the performance of our proposed models. highlight how observed confounders—identified through causal discovery techniques—can influence causal associations, we make use of a synthetic dataset. In addition, to evaluate the overall performance of our proposed model structures, two different types of datasets with exclusive properties, a Virtual Reality (VR) based Stated Preference (SP) dataset and a Revealed Preference (RP) dataset, are used. The VR dataset is collected by immersing the participants in artificial computer-generated experiences \parencite{farooq2018virtual}. An openly available RP dataset, the London Travel Demand Survey (LTDS), to investigate travel demand in the metropolitan area of London is also used \parencite{hillel2018recreating}.

The rest of this paper is organized as follows. First, a brief review of the literature on causality analysis and counterfactual modelling in transportation is provided in Section \ref{Background}. It is followed by Methodology (Section \ref{methods}) in which the structure of the two proposed models is described in detail. In Section \ref{data}, a description of datasets is presented and then the analysis of estimated models for SP and RP datasets and also synthetic datasets are provided in Section \ref{result}. Finally, in Conclusion (Section \ref{conclusion}), some suggestions and future directions for the study are proposed.

\section{Background} \label{Background}
In this section, first, a brief overview of causality analysis in transportation studies is presented. Secondly, deep leaning based counterfactual modelling applications in transportation are discussed.

\subsection{Causality in Travel Behaviour Analysis}
%Behavioural models are primarily utilized in policy assessment, where the focus is on answering what-if questions. These causality-driven questions aim to understand how changes in a specific factor influence the behaviour of a system. 
In the recent years, transportation researchers have increasingly emphasized on causal analysis, particularly for addressing the issue of self-selection and endogeneity in travel behaviour studies \parencite{mokhtarian2008examining,cao2009examining, golob2003structural, vance2007impact, kamal2022debiased, brathwaite2018causal}. The self-selection issue in travel behaviour modelling arises when individual's travel choices are influenced by their inherent attitudes, preferences or sociodemographic characteristics rather than the direct impact of a variable of interest or a policy. %For instance, individuals with lower incomes or without car ownership might opt for residences in areas with available public transit options, leading to higher transit use. In fact, travel mode choices may be shaped by individuals' underlying attitudes or sociodemographic characteristics, rather than being solely a response to environmental factors.
Without properly accounting for this issue, models may incorrectly attribute changes in travel choices to the policy, when, in fact, those changes are largely driven by individual's unobserved or observed characteristics. In fact, the self-selection problem blurs the distinction between correlation and causation between two variables. In terms of unobserved characteristics, this issue is also referred to as endogeneity bias, where explanatory variables are correlated to unobserved variables either directly or through common causes, resulting in a biased estimation of policy impact.

\begin{figure}[h]
\centering
    \includegraphics[width=0.8\textwidth]{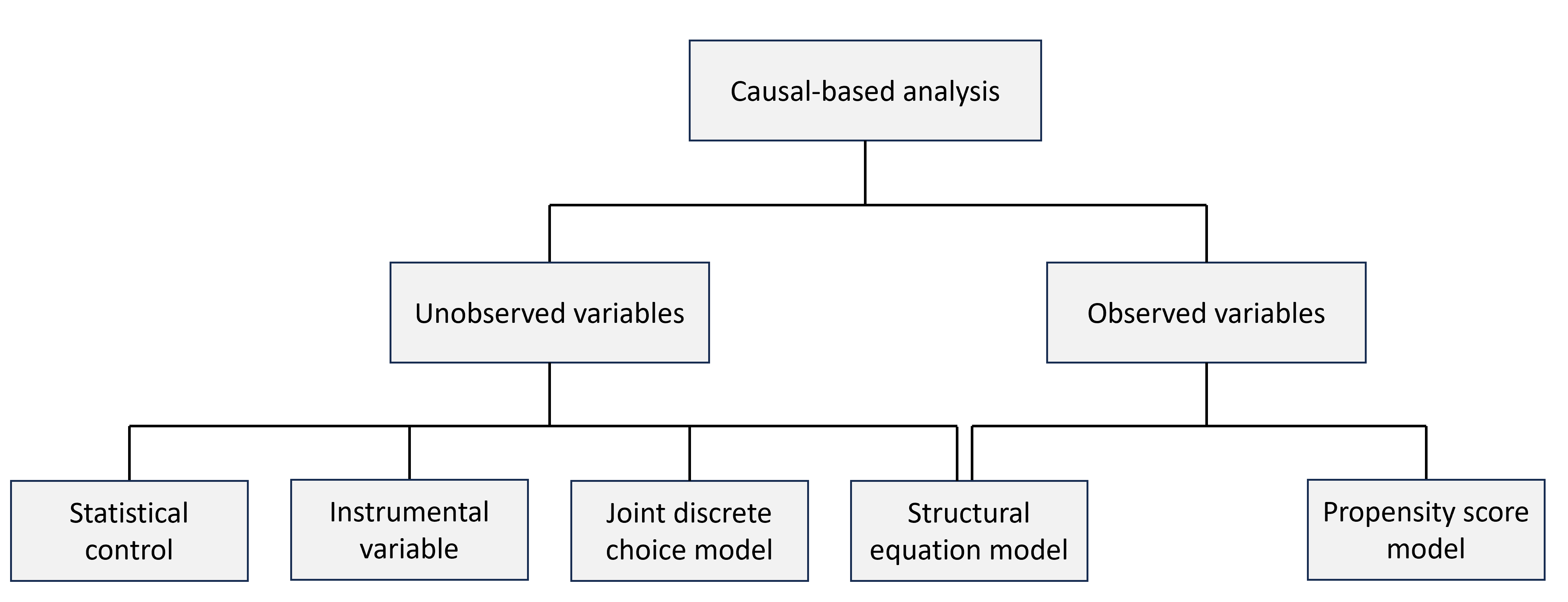}
  \caption{Causal analysis classification based on observed and unobserved confounders}\label{fig: causal models classification}
\end{figure}

Using observational datasets, various approaches such as statistical control, instrumental variables, propensity score, joint discrete choice models and structural equations models have been employed in transportation studies. Figure \ref{fig: causal models classification} demonstrates the classification of these methods based on their focus on observed and unobserved confounders.

Some studies statistically control the endogeneity problem by explicitly transforming unobserved variables into measurable and observed ones \parencite{kitamura1997micro,handy2005correlation, cao2006influences}. Including unobserved variables in the model allows modellers to distinguish between effects driven by attitudes and those influenced by policy interventions. However, accurately measuring attitudes is challenging, as they are often prone to measurement errors. {Inspired by this statistical approach,} \cite{kamalFarooq_2025} {proposed Copula-ResLogit, a framework that integrates Residual Neural Networks (ResNet) various copulas to account for unobserved confounding. It first detects hidden, non-causal associations using traditional copula functions and then leverages deep residual layers to mitigate these confounding effects. The proposed framework effectively disentangle the effects of unobserved factors from true causal relationships, while maintaining model interpretability. We therefore use Copula-ResLogit to handle any unobserved conformity.}

The Instrumental Variables (IV) method is another common approach addressing the endogeneity problem in the opposite way. 
%This approach first discovers variables, called instruments, that are minimally correlated with error terms. The instruments are used to predict the value of the variables of interest (policy) and then the predicted value of the policy is fed into the model to estimate an outcome \parencite{vance2007impact}. In fact, the instruments help isolate the variation in the explanatory variables that are not correlated with the error term. 
The main challenge that researchers encounter with this approach is ``weak instruments'' \parencite{bound1995problems,cameron2005microeconometrics}. Instruments must be strongly correlated with the policy while remaining uncorrelated with the error term in the outcome equation. In addition, the instruments must not have a direct impact on the outcome. Failing to meet either of these criteria can result in biased estimates.
The propensity score method, classified under the potential outcome frameworks \parencite{rubin2005causal}, is commonly applied in various fields to address the issue of self-selection rooted in observed characteristics. 
%By applying adjustment techniques such as regression, matching, and stratification, this approach seeks to mimic the conditions of a randomized experiment \parencite{cao2012exploring, cao2010exploring}. 
The causal effect is defined as the comparison between two potential outcomes. {However, as both potential outcomes are not simultaneously observed for each individual}, modellers strive to provide comparability in observational studies by establishing a balance between the treatment and control groups that differ in their exposure to the policy. The propensity score is used to balance the selection process based on observed characteristics, enabling us to attribute any differences in behaviour to the variable of interest (policy).

%Sample selection model is another approach that models the outcome of interest as conditional on a prior selection process in travel behaviour. The concept behind this method is that the observed decision or outcome in the dataset is not random but is influenced by an underlying selection process. Sample selection models generally involve two modelling steps. 1) Selection equation: initially, the individual's propensity or the process of an individual’s selection is modelled, 2) Outcome equation: the second step involves predicting and analyzing the individual’s specific travel behaviour. 
In discrete choice modelling, joint models can be classified under the sample selection technique, where the behavioural process is decomposed into separate components (equations) but estimated jointly in one step through full-information estimation \parencite{zhou2008self,bhat2007comprehensive}. This model structure captures the correlation between unobserved variables of multiple endogenous choices, representing confounding effects rooted in common unobserved variables. Conventionally, the normal distribution was assumed for the correlation between errors of equations. However, the copula approach has empowered researchers to capture a stochastic correlation among variables without being restricted by specific probability distributions. \parencite{bhat2009copula, rana2010copula,keya2019joint}. Structural Equation Model (SEM) is also categorized under simultaneous models in the travel behaviour modelling \parencite{golob2003structural, golob1998greenhouse,wang2017influencing, aditjandra2012understanding, cao2007changes,fujii2000anticipated}. SEM is a modelling technique that can capture the interaction between a large number of endogenous, exogenous and unobserved variables influencing the behaviour-generating process. SEM is a specific type of structural causal model (SCM), but unlike SEM, in SCM, the associations between endogenous and exogenous variables are generally modelled using nonlinear or nonparametric functions \parencite{spirtes2001causation, pearl2000models, maat2009causal}.

\cite{brathwaite2018causal} discussed the fundamental problem of the lack of using Directed Acyclic Graphs (DAGs)-based approaches in advanced travel behaviour models and highlighted the essential requirement of applying both causal inference and machine learning algorithms in behavioural analysis. This study highlighted Pearl's DAGs approach in which causal identification is prior to estimation \parencite{pearl2009causality}. In observational studies, this method of causal discovery involves 1) causal graph identification and 2) converting causality quantity to a statistical expression. 
{Causal discovery is a set of algorithmic techniques used to derive causal graphical models, such as DAGs, from observational, interventional, or experimental data} \parencite{shen2020challenges}. {For details on the definition and properties of DAG, please refer to Appendix} \ref{app:AA}. Inspired by \cite{brathwaite2018causal}, our study endeavour to initiate a new research area in causality analysis within transportation utilizing the DAG-based approach and deep learning algorithms. 

\subsection{Counterfactual Modelling with Deep Learning}
\label{Counterfactual backgrounf}
{In the context of time-varying treatments and traffic/flow forecasting, several studies generate counterfactual predictions by learning causal graph from observed input features and leveraging deep learning algorithms. For example, the Counterfactual Graph Transformer (CGT) integrates counterfactual explanation with graph transformer models to improve interpretability and prediction accuracy in traffic flow forecasting \parencite{yang2023counterfactual}. The key idea is to identify which spatial graph connections and temporal input features truly drive predictions. To this end, it uses a perturbation mask generator over sensor features to obtain spatial and temporal counterfactual predictions. \cite{deng2023spatio} proposed a Neural Structural Causal Model (STNSCM) for bike flow prediction that integrates causal reasoning with deep learning. Their approach models nonlinear spatio-temporal dynamics using graph convolution and gated neural structures, applies the front-door criterion to mitigate confounding bias, and includes a counterfactual reasoning module to infer spatio-temporal states under counterfactual scenarios. Another study by \cite{li2024msct} proposed Marginal Structural Causal Transformer (MSCT) to predict counterfactual post-crash traffic conditions by combining Marginal Structural Models, designed for time-dependent treatments, with a Transformer-based sequence model, enabling traffic-speed prediction under hypothetical crash interventions while correcting for time-varying confounding. Typically, in the existing transportation literature, deep learning is directly applied in order to model the causal mechanisms inside the graph.

{Beyond transportation literature, several examples can be found where structural causal models for counterfactual inference are developed using deep learning.} \cite{pearl2009causal} {introduced the do-calculus and counterfactual logic with a three step approach: Abduction $\rightarrow$ Action $\rightarrow$ Prediction. However, the Abduction step was mathematically intractable for complex data, e.g., travel behaviour surveys, especially in terms of recovering unobserved noise variables.} \cite{louizos2017causal} {proposed using Variational Autoencoders (VAE) to recover the unobserved noise variables from the proxy variables and allowing for an unbiased Individual Treatment Effect (ITE) estimation. However, the method relies on the Evidence Lower Bound approximation rather than an exact inverse, resulting in probabilistic counterfactuals.} \cite{pawlowski2020deep} {solved this issue by using Normalizing Flows and VAE together to invert the causal mechanisms and deterministically recover the noise.} \cite{kocaoglu2017experimental} {used Generative Adversarial Networks (GANs) to develop the noise recovery, but it required running an optimization loop for every single individual at the time of inference.} \cite{sanchez2022diffusion} {used the diffusion model to solve the same problem, but it required dealing with ordinary differential equations, which is computationally demanding. It is noted that among these methods, in the context of travel behaviour analysis, Normalizing Flow appears to be the most promising candidate. It is deterministic and, with a complexity of $O(1)$, only requires one forward pass to perform the Abduction step and one forward pass to perform prediction.}

To the best of our knowledge, an interpretable hybrid causal framework for behaviour modelling remains missing from existing transportation literature. We leverage econometrics, deep learning, and recent advancements in structural causal models for counterfactual inference to: 1) mitigate the impact of unobserved confounders within the interpretable structure of {Random Utility Maximization (RUM)} theory and 2) learn latent representations that support counterfactual modelling and forecasting.}

\section{Methodology}
\label{methods}
 Within \emph{deep CAusal infeRence mOdelling for traveL behavIour aNAlysis (CAROLINA)}, we propose two model structures: 1) interpretable deep structural causal model, and 2) generative counterfactual model. The first framework tries to isolate the direct (causal) impact of a variable of interest on an outcome from observed and unobserved confounding associations. Furthermore, the second proposed framework presents individual-level counterfactual analysis, enabling transportation planners to predict outcomes in a manipulated system based on causality analysis.

\subsection{Interpretable Deep Structural Causal Model}
\label{IDSCM}
\begin{figure}[!ht]
  \begin{center}
   \includegraphics[width=1.025\textwidth]{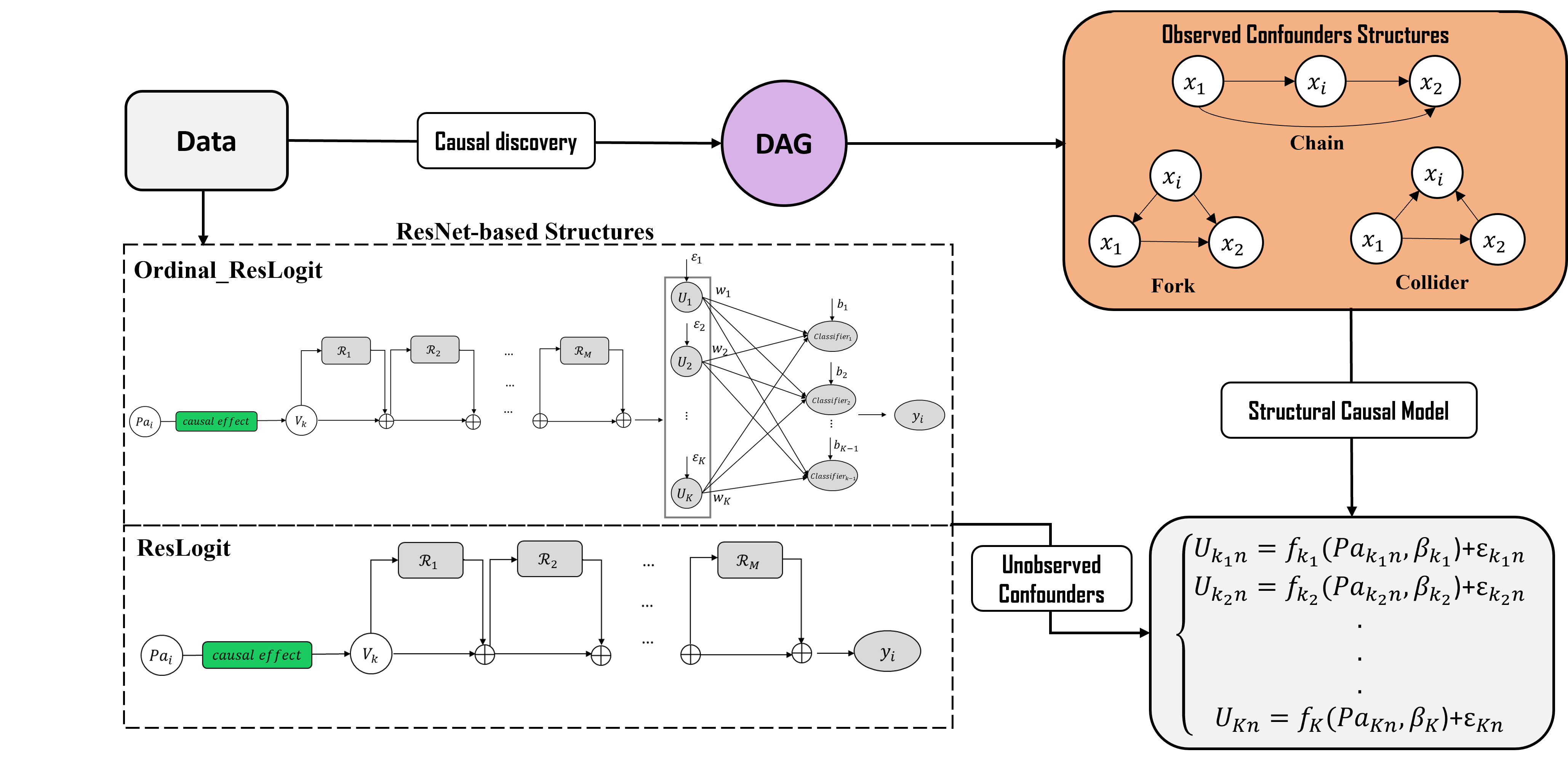}
   \caption{The architecture of interpertable deep structural causal model in \emph{CAROLINA}}\label{DSCMframework}
   \end{center}
\end{figure}

The main objective of this framework is to overcome the impact of both observed and unobserved confounding variables. We simultaneously isolate the causal association from the impact of these two types of confounders but with different approaches. To that end, our study deals with unobserved non-causal associations by employing the statistical control method but using a deep learning-based component. The power of this approach was validated through the copula approach in our recent study \parencite{kamalFarooq_2025}. {The study proposed a hybrid data-driven copula-based joint behavioural model that investigates the ability of deep learning algorithms to disentangle true causal relationships from non-causal associations induced by unobserved confounding factors. It consists of two main parts: first it explores the correlation between dependent variables in the choice selection process caused by unobserved factors. Secondly, it examines the power of deep learning approaches to mitigate associations arising from unobserved confounding in support of causality-based decision-making. {In this paper, we focus on unobserved variables whose contribution is represented through the structural error and assume that this unobserved component is independent of the observed exogenous variables. In other words, we overcome the unobserved non-causal association between treatment and outcome which is rooted in their structural errors}. The study demonstrated that statistically control unobserved confounders through deep learning component allows modellers to effectively eliminate those unobserved dependencies.} Furthermore, with respect to observed confounders, our proposed model strives to isolate the causal association from observed confounding impacts. This aspect of our model leverages causal discovery techniques to infer the underlying causal structure and identify potential non-causal associations arising from observed confounders. The structure of the interpretable deep structural causal model can be found in Figure \ref{DSCMframework}.

The first step of our proposed model is to identify the causal structure of a system and observed confounders. This step focuses on non-parametric identification rather than estimation. We employ causal discovery algorithms to identify DAGs from observational datasets, while incorporating a certain level of domain knowledge. This prior knowledge can help modellers in preventing two types of spurious edges. First, it ensures that the edges among some variables, such as socio-demographic variables are prohibited. Second, in temporal studies, it restricts edges from variables representing later time points to those at earlier time points \parencite{shen2020challenges}. Generally, causal discovery is a powerful approach that allows us to infer causal relationships from observational data without prior assumptions or domain knowledge. However, in order to improve the causal discovery process and reduce the risk of identification issues in DAG learning, commonly known as the Markov equivalence class problem, we incorporate the domain-specific background knowledge. Two graphs are considered Markov equivalent if they correspond to the same set of conditional independence relationships. Incorporating background knowledge helps narrow the search space of possible DAGs, reducing the potential identification problem in causal discovery algorithms. {Among the listed algorithms in Appendix} \ref{app:AA}{, here we used FGES as it treats the discovery as a holistic optimization problem, is fast score-based, more robust to non-faithfulness, highly scalable to large and complex datasets, and exhibits the properties of consistency, efficiency, and stability.}

\begin{figure}[!ht]
  \begin{center}
   \includegraphics[width=0.8\textwidth]{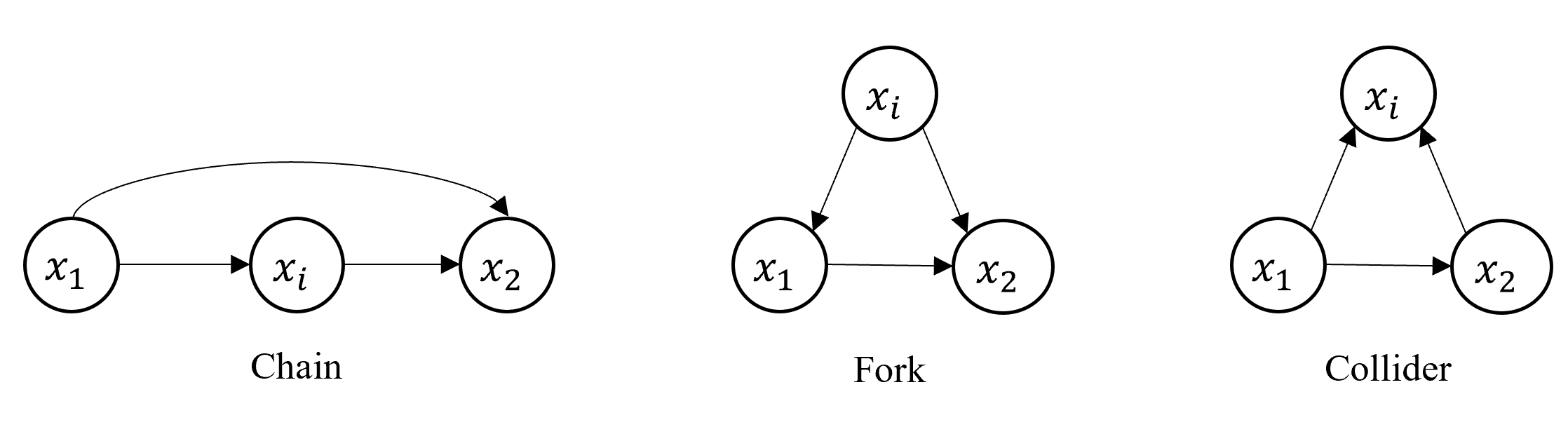}
   \caption{Three basic structures in a DAG}\label{basic structures}
   \end{center}
\end{figure}

{The obtained DAG topology represents qualitative causal orderings among observed factors} and enables us to infer the behaviour-generating process and observed confounding variables. It is important to note that, at the identification step, we assume no unobserved confounders affect the DAGs and in the next step, we address unobserved confounding variables by using the statistical control method and leveraging the advantages of deep learning techniques \parencite{kamalFarooq_2025}. {More specifically, we focus on unobserved variables that induce dependence between the structural disturbances of the dependent variables. We assume that this unobserved component is independent of the observed exogenous variables used in the causal discovery stage}. In our proposed framework, the DAG provides valuable information about the variable selection of utility functions in discrete choice modelling. In other words, we can analyze how conditioning on a variable can either create or eliminate non-causal associations. There are three main structures in each DAG illustrated in Figure \ref{basic structures} \parencite{pearl2009causality}.

These structures are interpreted as follows: 

\begin{itemize}
    \item Chain structure ($x_{1}\rightarrow x_{i}\rightarrow x_{2}$ or $x_{1}\leftarrow x_{i}\leftarrow x_{2}$): The indirect (non-causal) relationship between $x_{1}$ and $x_{2}$ is blocked, if we condition on $x_i$.
    \item Fork structure ($x_{1}\leftarrow x_{i}\rightarrow x_{2}$): The indirect (non-causal) relationship between $x_{1}$ and $x_{2}$ is blocked, if we condition on $x_i$.
    \item Collider structure ($x_{1}\rightarrow x_{i}\leftarrow x_{2}$): The indirect (non-causal) relationship between $x_{1}$ and $x_{2}$ is unblocked, if we condition on $x_i$.
\end{itemize}

To deconfound the causal association between the variables of interest from observed confounders, our approach employs the back-door identification strategy \parencite{pearl2009causality}. This method ensures that all non-causal associations are blocked while preventing the creation of new spurious paths. Based on DAG structures, we need to: (1) condition on a non-collider set and (2) avoid conditioning on a collider set. The Structural Causal Model (SCM) is a type of causal inference framework that relaxes non-parametric analysis and models the underlying behaviour-generating process identified in DAGs \parencite{pearl2009causality}. The SCM is logically equivalent to the potential outcomes framework; however, it enhances the adjustment approach for handling confounders by considering a non-parametric graphical model prior to estimation. The structural causal model accounts for the impact of observed confounders based on the back-door criteria, resulting in eliminating non-causal associations. This model structure estimates how each variable in the system interacts with its causes (parents). Mathematically, each SCM is a single model that consists of a collection of asymmetric structural equations:

\begin{equation}
\label{SCM-equetion}
\begin{split}
x_i:=f_i(Pa_i, \beta_i) + \varepsilon_i
\end{split}
\end{equation}

\noindent where $f_i$ represents structural mechanisms that generate an endogenous variable based on causes (parents) $(Pa_i)$. In addition, $\varepsilon_i$ is an exogenous latent variable representing sources of variation or noise in the modelling. Estimation of the SCM involves estimating the parameters of each structural causal mechanism $\beta_i$. Note that equality in equations represents symmetric; however, in the causality analysis, we put emphasis on asymmetrical equations. In DAGs, random variables $(x_{i})$ can function solely as an exogenous variable $(x_{ex})$, an endogenous variable $(x_{en})$ or serve the dual role of both. {While, the edge $\zeta_{ji}$ represents the causal relationship between $x_j$ and $x_i$}. To build the structural causal model, we first need to define the parents of all random variables.

\begin{equation}
    Pa_i =
    \begin{cases}
    \displaystyle
        \emptyset, &\text{if }\;\;  \displaystyle \sum_{{j\neq i}}^n \mathbb{I}(\zeta_{ji}) = 0 \\
        \{x_{j},\dots,x_{n}\}, &\text{if }\;\; \forall x_{j} \setminus \{x_{i}\},  \mathbb{I}(\zeta_{ji}) \neq  0
    \end{cases}
\end{equation}

\noindent where, for each $x_{i}$ that $Pa_i\neq \emptyset$ and $Pa_i = \{x_{ex}\} $, $x_{i}$ is referred to as the root of SCM. In the case of DCMs, for each individual $n$, we start by computing the utility function of $U_{kn}$ for $x_i$ with $K$ discrete alternative based on causal structures. In the following equations, for simplicity, $i$ is eliminated from notations.

\begin{equation}
    \begin{aligned}
        U_{k_{1}n} &=:f_{k_{1}}(Pa_{k_{1}n}, \beta_{k_{1}}) + \varepsilon_{k_{1}n}  \\
        U_{k_{2}n} &=:f_{k_{2}}(Pa_{k_{2}n}, \beta_{k_{2}}) + \varepsilon_{k_{2}n} \\
        &\;\; \vdots\\
        U_{Kn} &=:f_{K}(Pa_{Kn}, \beta_{K}) + \varepsilon_{Kn} 
    \end{aligned}
\end{equation}

To simultaneously deal with unobserved confounders, we employ a statistical control approach that transforms unobserved confounders into observed ones and integrates them into the model structure. {These unobserved confounders are common to the error terms of the dependent variables and introduce a non-causal association between them}. In our approach, we leverage the capability of deep neural layers to capture and measure unobserved confounders. This methodology builds on our recent study, which underscores the strength of deep learning components and validates our proposed control approach for common unobserved confounders through a joint modelling approach \parencite{kamalFarooq_2025}. {It mainly examine the power of deep learning to control for unobserved confounding association.} To account for unobserved confounders, the \emph{CAROLINA} employs ResNet architectures and thus serves as a logical extension of ResLogit and Ordinal-ResLogit, while being explicitly designed for causality analysis \parencite{wong2021reslogit,kamal2024ordinal}. These ResNet-based models utilize both linear and deep components in the utility function of discrete choice models (DCMs). Therefore, our proposed approach is able to control for common unobserved confounders between structural mechanisms while preserving the interpretability of observed variables. It is of note that \emph{CAROLINA} is flexible enough to use other machine learning structures, provided their performance in controlling unobserved confounders is validated based on \parencite{kamalFarooq_2025} study.

In ResNet-based models, to obtain the utility function of alternatives, first, a deterministic $K\times 1$ utility vector is computed, where each element represents the linear impact of causes or parents $(\beta_k Pa_{kn})$. This vector is then passed through $m$ residual layers using a $K\times K$ matrix of residual parameters $W^m$, resulting in the $K\times 1$ utility vector $(U_{kn})$. In this vector, each element represents the effect of $Pa_{kn}$ on the $k^{th}$ alternative of $x_i$ while also measuring and including unobserved confounders in the utility function. Mathematically, the utility function of each alternative $k$ for individual $n$ can be obtained from Equation \ref{utility function}. 
 
\begin{equation}
\label{utility function}
U_{kn}=:\beta_k Pa_{kn} + g_{kn} + \varepsilon_{kn} 
\end{equation}

In this equation, the deep residual component $(g_{kn})$ is the $k^{th}$ element of vector $g_{n}$ as follows:

\begin{equation}
\label{g_{kn}}
g_{n} = \displaystyle - \sum_{m=1}^{M} \ln \Bigg(1+ \exp \Big(W^m {V_n}^{(m-1)} \Big)\Bigg)
\end{equation}

The residual connection comes from the previous layer and ${V_n}^{(m-1)}$ is the output vector of non-linear utility components for the ${(m-1)}^{th}$ residual layer consisting of $K$ members:

\begin{equation}
\label{V_n}
{V_n}^{m}= {V_n}^{m-1} - \ln \Bigg(1+ \exp \Big(W^m {V_n}^{m-1}\Big) \Bigg)
\end{equation}

\noindent where ${V_n}^{0}$ is the vector of linear function of parents. Then, according to the obtained utility function, for each random variable $(x_{i})$, the probability of choosing each alternative of categorical or ordinal random variables is obtained as follows:

%\frac {\exp{(U_{kn})}} {\exp{(\sum_{k=1}^{K} U_{kn})}}

\begin{equation} \label{Prob}
    P(x_i=k | Pa_{k}) =
    \begin{cases}
     \displaystyle
         \frac{\exp{(\beta_k Pa_{k} + g_{k})}} {\displaystyle \sum_{k=1}^{K} \exp{\Big(\beta_k Pa_{k} + g_{k}\Big)}} , & \text{if }  x_{i} :\text{Categorical} \\
         \\
          \displaystyle\sigma \Bigg(\Big(\sum_{k=1}^{K} W_k (\beta_k Pa_{k} + g_{k})\Big) + b_k\Bigg) & \text{if } x_{i} :\text{Ordinal}
    \end{cases}
\end{equation}

\noindent where $W_k$ and $b_k$ are the weight and independent bias parameters of the penultimate layer of Ordinal-ResLogit in which the problem is reduced to $K-1$ binary classification problems. For further details on the ResLogit and Ordinal-ResLogit models, we direct the readers to \cite{kamal2024ordinal} and \cite{wong2021reslogit}. It is of note that in the modelling process, each exogenous variable can serve as a parent to another exogenous variable, represented as  $Pa^{x_{j}} = \{x_{ex},x_{i}\}$. This computation is subsequently continued until we reach the outcome variable or the variable playing only as a child or descendent $(x_{des})$ of other variables in the causal structure. This algorithm is mathematically presented in Appendix \ref{pseudo}. For training the model, the joint probability distribution is defined based on the Local Markov Property and Markov Factorization property as follows:

\begin{definition}[Local Markov property]
A random variable $x_{i}$ is independent of all its non-descendants given its parents, meaning that once we have information about the parents of variables, the variable is conditionally independent of all other random variables in the DAG that are not its descendants.
\end{definition}

\begin{definition}[Markov Factorization property]
Given a probability distribution $P$ and a directed acyclic graph  ${\mathcal{G}}$, $P$ is factorized according to ${\mathcal{G}}$:

\begin{equation}
\label{Markov factorization propertyy}
P(x_1,...x_I)=  \prod_{i=1}^{I} P\Bigg(x_i | Pa_i\Bigg) 
\end{equation}
\end{definition}

\noindent where for each causal mechanism, $P(x_i|Pa_i)$ is calculated using Equation \ref{Prob}. In our proposed model, in order to estimate the effect of parents $(\beta)$ and total deep layers parameters $(\theta)$ including residual layers and the penultimate layer of Ordinal-ResLogit, all causal mechanisms are jointly modelled. The loss function, as shown in Equation \ref{loss function-DSCM}, is minimized using a minibatch data-driven stochastic gradient descent (SGD) learning algorithm with an RMSprop optimization step \parencite{goodfellow2016deep}.

\begin{equation}
\label{loss function-DSCM}
LL(\beta,\theta)  = \sum_{n=1}^{N}\sum_{i=1}^{I} \Bigg[\ln (P(x_1,...x_I)M_{n}\Bigg]
\end{equation}

\noindent where $M_{n}\in \{0,1\}$ is the chosen alternative among joint alternatives.

\subsection{Counterfactual Model Structure}
\begin{figure}[!ht]
  \begin{center}
   \includegraphics[width=0.8 \textwidth]{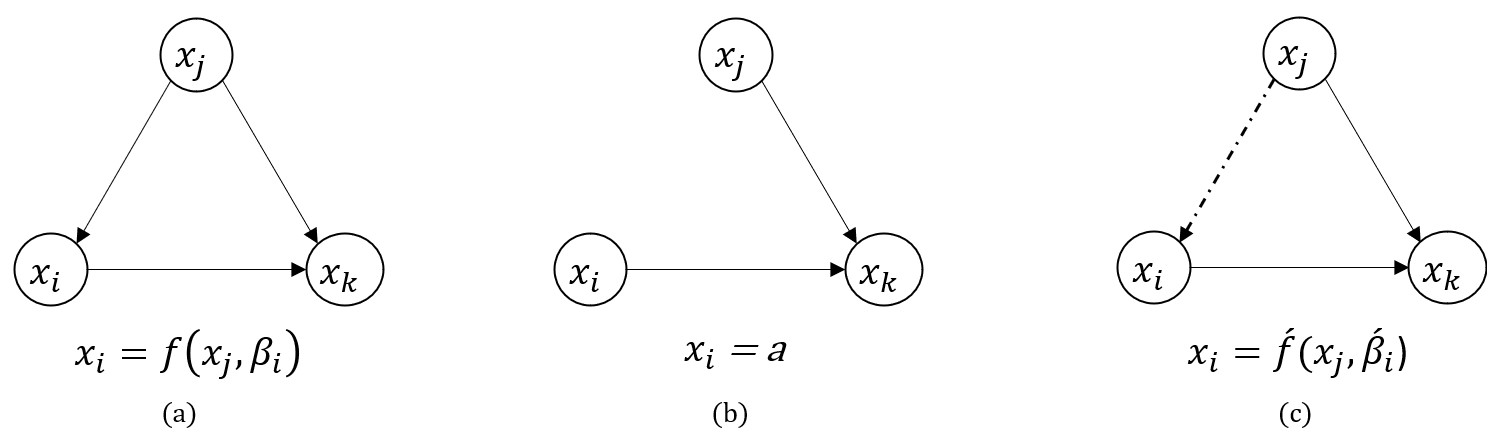}
   \caption{Visualization of Intervention Operations (a): original, (b): hard, (c): soft}\label{intervention}
   \end{center}
\end{figure}

 A counterfactual distribution, mathematically denoted by the do operator $P(Y|do(x_{i}))$, refers to the probability distribution of an outcome after externally intervening in a random variable of interest $x_{i}$, known as treatment or policy \parencite {pearl2009causality, pearl2009causal, shpitser2006identification}. The counterfactual world is conceptually different from the observational distribution. In causal analysis, SCMs allow the study of counterfactual scenarios in which various interventions or policies may occur. The intervention operation ($do$ operator) can be hard or soft as shown in Figure \ref{intervention}, depending on whether a variable is set to a specific value $x_{i}:= a$, or whether a new structural mechanism is defined $x_{i}:= f^{'}(Pa_i, \beta_i) + \varepsilon_i$. In other words, in the causal graph, the $do$ operator means that all incoming links on the node variable of interest should be removed or functionality changed, and accordingly the corresponding causal mechanism should be changed in the SCM. Therefore, in the manipulated system, the $do$ operator graphically changes the causal structures in a directed acyclic graph. Notably, the $do$ operation on the variable $x_{i}$ only changes the causal mechanism of $x_{i}$ and does not change other parts of the behaviour-generating process. This fact is rooted in the Modularity assumption or the Law of Counterfactuals, stating that in the structural causal model, causal mechanisms are modular or independent \parencite{pearl2009causality}.

An interventional structural causal model or counterfactual model represents intentional manipulations in a data-generating process. This model is built based on the current causal structure, where some causal mechanisms are changed, but the latent variables remain unchanged \parencite{pearl2009causality, peters2017elements}. Counterfactual analysis is a query at the unit level and allows us to generate counterfactual outcomes under different intervention scenarios and compare them with observed outcomes. In other words, our counterfactual model structure assists us in making inferences about how individuals would change their decisions if some interventions were implemented in the system. Our proposed model allows us to compute both the average and individual effects of an intervention (treatment). A typical counterfactual modelling process consists of three levels: abduction, manipulation, and prediction \parencite{pearl2009causality}. These steps are clarified and visualized in Figure \ref{steps of counterfactual models} in Appendix \ref{app:B}. The figure also compares the counterfactual prediction model with the classic forecasting approach. In contrast to the association-based prediction model, our proposed model initially tries to eliminate the impact of confounders, and subsequently, upon learning the latent structure, the model predicts the counterfactual outcome by considering manipulations in the causal mechanisms.

\begin{figure}[!ht]
  \begin{center}
   \includegraphics[width=0.6\textwidth]{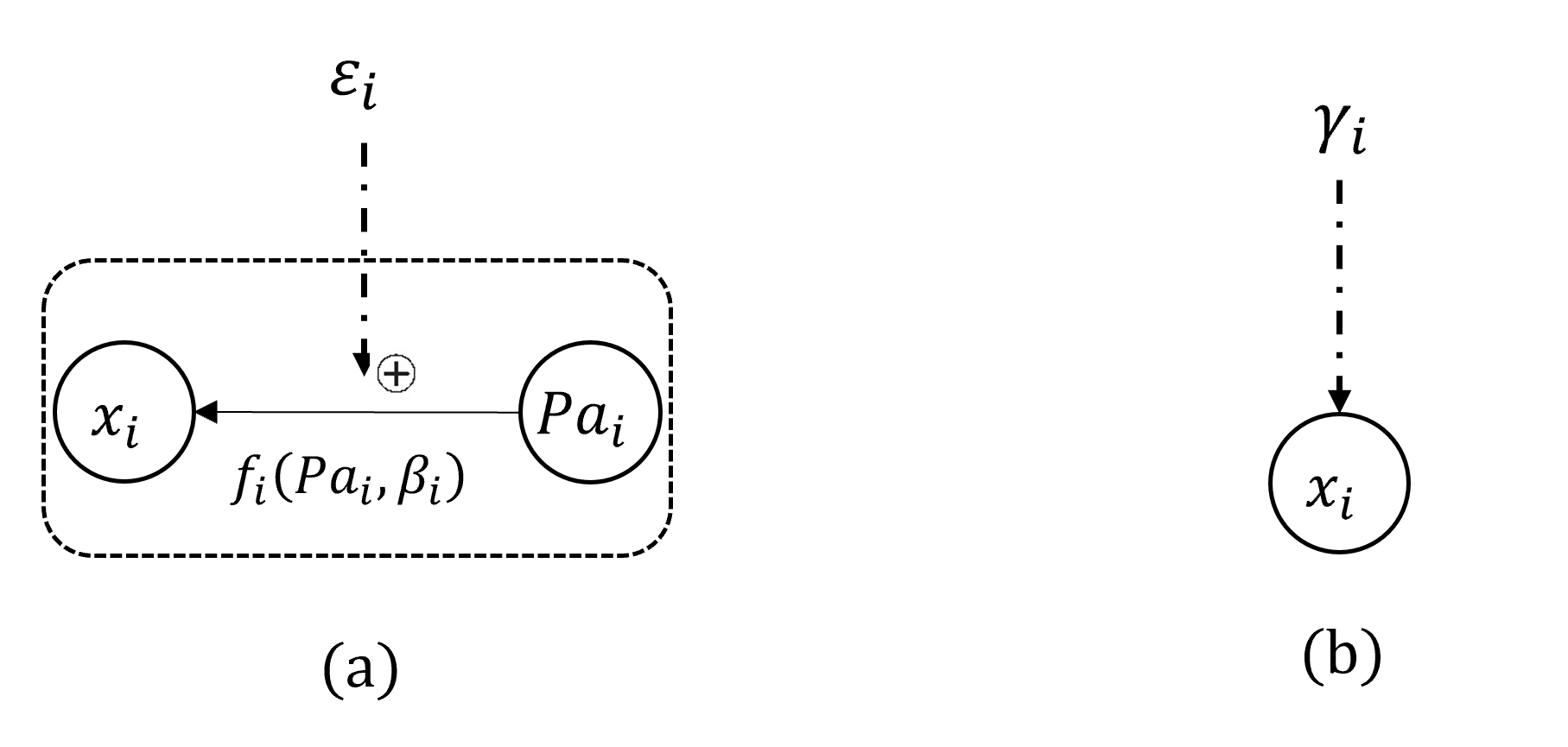}
   \caption{The visualization of latent construct in a DAG}\label{example}
   \end{center}
\end{figure}

The abduction step is the first step of the counterfactual model for inferring the latent exogenous variables of each causal mechanism $\varepsilon_i$. This step helps us to understand the uncertainty and variability in the system that is assumed to be identical between the factual and counterfactual world. To clarify this, let us focus on one part of an obtained DAG shown in Figure \ref{example}a. In this example, $Pa^{x_i} = \{Pa^{i}_1, Pa^{i}_2, ..., Pa^{i}_n\}$ is the set of parents of a discrete binary random variable $x_i$ and for this part of DAG the utility function of this variable is $U_i=:f_i(Pa_i, \beta_i) + \varepsilon_i$. At the abduction level, we aim to approximate $\varepsilon_i$ which is a part of generating the endogenous variable. This variable is known as the latent or noise variable.

To estimate $\varepsilon_i$, we assume another form of causal mechanism generating process that only depends on the compressed latent representation of the observed variables, denoted as $\gamma_i$ in Figure \ref{example} (Figure (b)). To capture a more accurate approximation of $\varepsilon_i$, we leverage the capabilities of both the Variational Auto Encoder (VAE) and normalizing flow. In the context of posterior distribution inference, VAE has been used in various disciplines to learn a latent space that captures the underlying factors of variation in the observed data \parencite{kingma2019introduction}. The VAE consists of an encoder network that maps the observed data to a latent space, and a decoder network that reconstructs the data from the encoded compressed latent representation of the observed variables. In other words, in our study, the encoder network is able to parameterizes the approximate posterior distribution $P(\gamma_i|pa_i)$, while the decoder network models the conditional distribution $P(pa_i|\gamma_i)$. Therefore, in the inference stage, given observed data $Pa_i$, the encoder network can be used to obtain a sample from the approximate posterior distribution $P(\gamma_i|Pa_i)$. The VAE traditionally assumes a simple distribution, such as a Gaussian distribution, for the latent space. However, this assumption may constrain the ability of a model to infer a flexible enough posterior distribution that accurately represents the underlying latent space of observed variables.

By incorporating the Normalizing Flow technique into the VAE, we can flexibly transform the simple distribution into a more representative distribution \parencite{kingma2016improved, pawlowski2020deep}. Generally, the normalizing flow model is a powerful generative modelling technique that allows for the flexible transformation of a simple base distribution to a more representative distribution \parencite{papamakarios2021normalizing, kobyzev2020normalizing}. In the literature, although the wide use of normalizing flows is to perform density estimation, flows-based models have emerged as a powerful tool for posterior approximations \parencite{papamakarios2021normalizing}.

Our Flow-based VAE framework involves learning the latent space representation of a causal mechanism denoted by $\gamma_i$. By comparing both structures in Figure \ref{example}, the $\varepsilon_i$ is computed as: 

\begin{equation}
\label{transformation}
\varepsilon_i = \gamma_i - f_i(Pa_i, \beta_i)
\end{equation}

The Flow-based VAE framework consists of three structures, encoder, normalizing flow model, and decoder. In our study, we use two MultiLayer Perceptron (MLP) structures for modelling the encoder and decoder of the VAE. First, based on causal mechanisms obtained from the DAG, the encoder structure takes observed data $(Pa_i)$ and generates the mean $(\mu)$ and standard deviation $(\sigma)$ of the initial variable of the flow model. The reparameterization process is used to set the simple standard Gaussian distribution as a base distribution of latent variable $\gamma_0 \sim \mathcal{N}(\gamma_0 | \mu,\sigma^2)$ \parencite{rezende2015variational}. The $\gamma_0$ value is then fed into the first flow layer. Normalizing flow structures consist of a stack of flow layers and transform the simple distribution into a distribution which is more representative of the true underlying distribution $(Q_k(\gamma_k))$. In fact, in the abduction level, both the encoder and normalizing flow model are used to infer latent constructs. Note that $\gamma_k$ is the latent space representation of observed variables of the causal mechanism and can be computed as:

\begin{equation}
\label{transformation}
\gamma_k = h_k \circ h_{k-1} \circ ... \circ h_1(\gamma_0)
\end{equation}

The distribution of the last variable equals the product of absolute values of determinants of Jacobin for each transformation \parencite{papamakarios2021normalizing, kobyzev2020normalizing, rezende2015variational}.

\begin{equation}
\label{distribution}
Q_k(\gamma_k) = Q_0(\gamma_0) \displaystyle \prod_{k=1}^{K} { \Bigg \vert \det \Bigg({\frac{\partial{h_k}}{\partial{\gamma_{k-1}}}}\Bigg)\Bigg \vert}^{-1}
\end{equation}

In Equation \ref{distribution}, the determinant term is computationally expensive. In order to control the efficiency of the model, in our study, we use simple planar flow \parencite{ kobyzev2020normalizing, rezende2015variational}. The formula for this transformation is written as follows:

\begin{equation}
\label{Planer}
h(\gamma) = \gamma + uh(w^{T}\gamma+b)
\end{equation}

\noindent where $u$, $w$ and $b$ are learnable parameters of the normalizing flow structures. Finally, the decoder takes the latent variable $(\gamma_k)$ and models the observed exogenous variables.

To train the Flow-based VAE model, we simultaneously estimate the set of parameters of encoder and normalizing flows $(\phi)$ and also decoder parameters $(\psi)$ by minimizing the negative lower bound of log-likelihood called Evidence Lower Bound (ELBO). ELBO  consists of two terms: 1) Kullback–Leibler(KL) divergence between the approximate posterior and the actual distribution, and 2) reconstruction error \parencite{rezende2015variational}. The negative ELBO can be computed as:

\begin{equation}
\label{Loss function 3}
-ELBO(\phi, \psi) =  KL \Big[ Q_{\phi}(\gamma|Pa) || P_{\psi}{(\gamma)}\Big] -\mathbb{E}_{Q_{\phi(\gamma|Pa)}}\Big[\ln \, P_{\psi}{(Pa|\gamma)}\Big]
\end{equation}

where the KL divergence is:
\begin{equation}
\label{Loss function 4}
KL \Big[ Q_{\phi}(\gamma|Pa) || P_{\psi}{(\gamma)}\Big] = \mathbb{E}_{Q_{\phi(\gamma|Pa)}}\Big[\ln \, Q_{\phi}(\gamma|Pa)\Big] -\mathbb{E}_{Q_{\phi(\gamma|Pa)}}\Big[\ln \, P_{\psi}{(\gamma)}\Big]
\end{equation}

Therefore, Equation \ref{Loss function 4} can be written as: 

\begin{equation}
\label{Loss function 5}
-ELBO(\phi, \psi) = \mathbb{E}_{Q_{\phi(\gamma|Pa)}}\Big[\ln \, Q_{\phi}{(\gamma|Pa)}- \ln \,P_{\psi}(Pa, \gamma)\Big]
\end{equation}

Given three components of the flow-based VAE model and parametrization technique, the negative ELBO is obtained as follows \parencite{rezende2015variational}:

\begin{equation}
\label{Loss function 6}
-ELBO(\phi, \psi) = \mathbb{E}_{Q_{0(\gamma_0)}}\Big[\ln \, Q_K{(\gamma_K)} - \ln \, P(Pa, \gamma_K)\Big]
\end{equation}

\noindent and finally, by substituting \ref{distribution}, the loss function can be computed as:  

\begin{equation}
\label{Loss function 7}
-ELBO(\phi, \psi) = \mathbb{E}_{Q_{0(\gamma_0)}}\Big[ \ln \,Q_0{(\gamma_0)}\Big] -\mathbb{E}_{Q_{0(\gamma_0)}}\Bigg[\sum_{k=1}^{K} \ln \, {\Bigg \vert det {\frac{\partial{h_k}}{\partial{\gamma_{k-1}}}\Bigg \vert}}\Bigg] -\mathbb{E}_{Q_{0(\gamma_0)}}\Big[ \ln \, P(Pa, \gamma_K)\Big]
\end{equation}

Minimizing this loss function with respect to $\phi$ and $\psi$ ensures the optimization of both the variational inference and generative components of the model. As observed in Equation \ref{Loss function 3}, a small KL divergence results in a tighter ELBO, which serves as a superior lower bound for $\ln \, P_{\psi}{(Pa_i)}$, resulting in unbiased generative model parameters. Furthermore, through the utilization of normalizing flow, we try to minimize KL divergence and enhance the flexibility of the encoder component in the VAE. Consequently, optimizing the Equation \ref{Loss function 7} not only ensures a reduction in KL divergence but also enhances the performance of the generative model.

After completing the abduction level and estimating the latent term of causal mechanisms $(\varepsilon_i)$, we need to develop the interventional SCM, representing the manipulated system. This step involves specifying intervention(s) or policy and manipulating the causal mechanism either in the soft or hard operation, shown in Figure \ref{intervention}. In this study, we manipulate the system based on the soft operation and we define the chosen intervention according to our case studies. Eventually, the interventional SCM or counterfactual model can be used to predict the counterfactual outcome and any changes in individuals' decisions after manipulation. The complete structure of our proposed counterfactual model consisting of abduction, manipulation, and prediction is illustrated in Figure \ref{counterfactualframework} and the algorithm of our proposed model can be found in Appendix \ref{pseudo}.
{It is noted that the measurement of unobserved confounding effects in our proposed models operates a bit as a black box, making it difficult for modellers to derive any meaningful behavioural insights from it. Moreover, this deep learning component may also include collider unobserved confounders and conditioning on them may potentially create new confounding pathways between variables of interest.}

\begin{figure}[h]
  \begin{center}
   \includegraphics[width=0.9\textwidth]{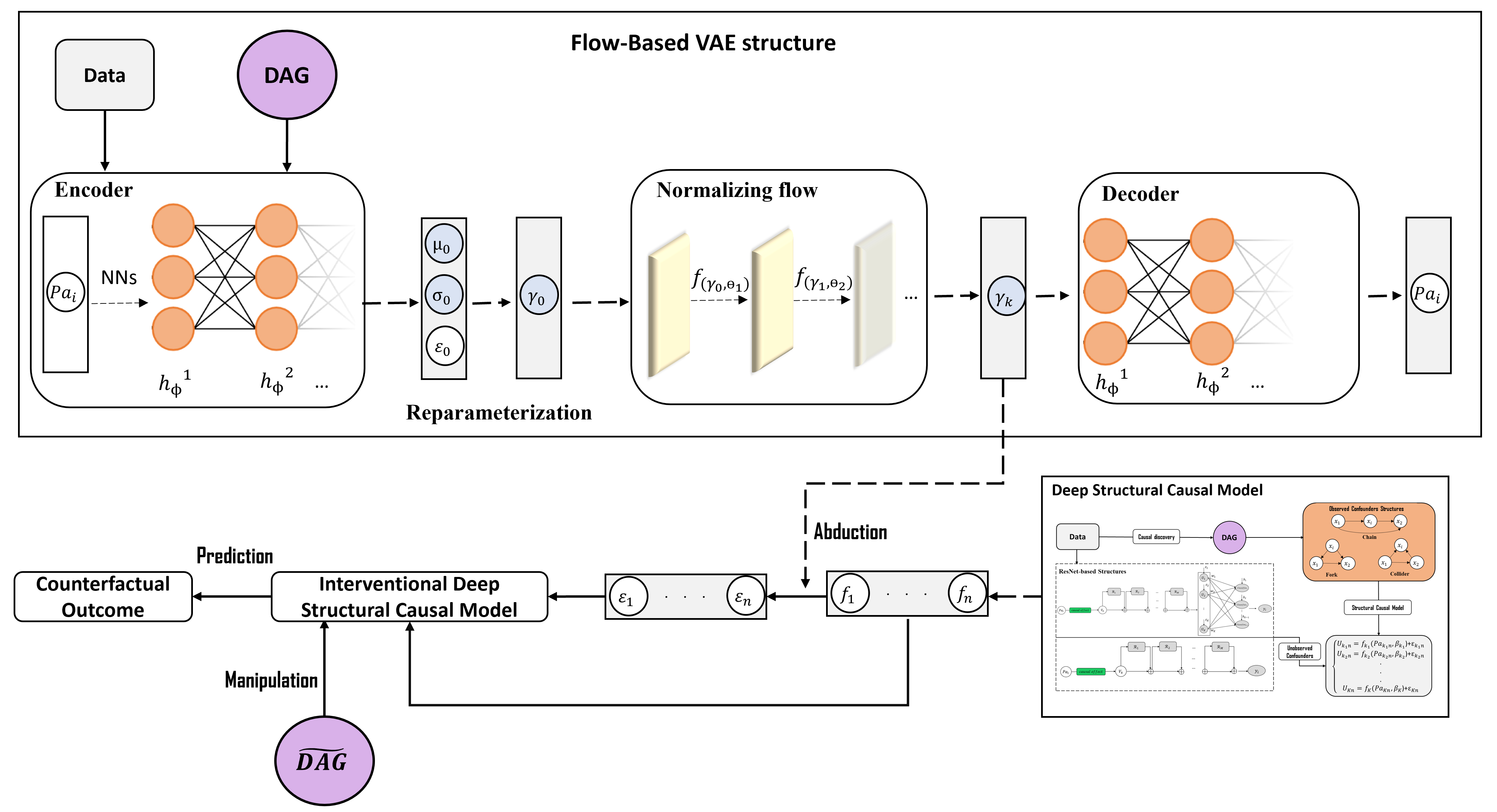}
   \caption{The architecture of generative counterfactual model in \emph{CAROLINA}}\label{counterfactualframework}
   \end{center}
\end{figure}

\section{Experiments on Simulated Dataset}
\begin{figure}[h]
  \begin{center}
   \includegraphics[width=0.4\textwidth]{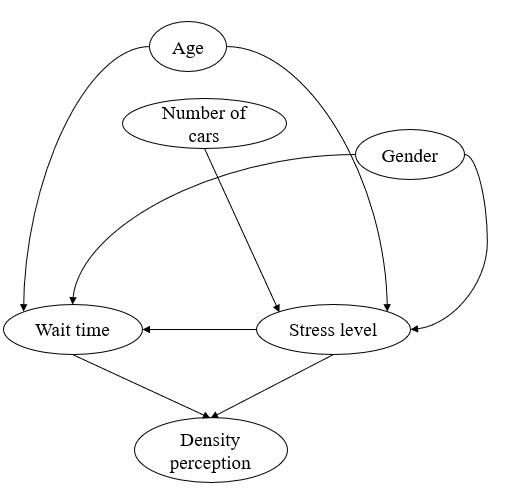}
   \caption{Causal structure of simulated dataset }\label{DAG-simulated}
   \end{center}
\end{figure}

Here we show the impact of observed confounders in causal analysis and highlight the difference between our proposed model and the typical travel behavioural modelling. Figure \Ref{DAG-simulated} shows the behaviour-generating process in the simulated dataset assumed by ourselves. In this graph, both collider: \{traffic density perception\} and non-collider: \{age, gender\} sets are assumed so as to clarify the role of these variables in the parameter estimation of policy (stress level) on the outcome (wait time). 

The generated synthetic dataset with 2,500 pseudo-observed individuals consists of five variables, including exogenous and endogenous variables. The exogenous variables are assumed to follow the same distributions as the observational pedestrian crossing dataset and the endogenous variables (stress level, waiting time, and density perception) are generated based on the ordered logit model assumptions with a set of known parameters. {Table \ref{simulateddataset} enumerates the detailed information of variables, their coefficients in the model and the assumed thresholds used in the ordered logit model specification.}  Note that the generated discrete density variable serves as a proxy for density perception in our simulated dataset. 

\begin{table}[H]
% \footnotesize
\scriptsize
\setlength{\tabcolsep}{4pt}
    \centering 
    \caption{{Variables of synthetic dataset}}\label{simulateddataset}
    \begin{tabular}{lcccccc}
    \hline
    %\addlinespace
    \textbf{Variables} & \textbf{Distribution} & \textbf{Categories} & \multicolumn{3}{c}{\textbf{ Coefficient value}}  & {\textbf{Thresholds}} \\
    %\addlinespace
    \hline
    & & & \textbf{Stress level} & \textbf{Wait time} & \textbf{Density perception}\\
    \hline
    Gender & Binary	& Yes, No	& 0.500	& 0.603	& - & -\\
    
    Age	& Categorical & Age 25-29 & 0.950 & -0.795& -& -\\
    & & Age 30-39 & 0.561 & -1.410& -& -\\
    & & Age 40-49 & 2.911 & -1.854& -& -\\
    & & Age over 50 & 1.975 & -2.854 & -& -\\
    
    Number of cars & Categorical & One car & -0.300 & -& -& - \\
    & & Over one car & -0.132 & -& -& - \\
    Stress level & Ordinal binary & High, Low & - & - & 0.500 & 1.05\\
    Wait time & Ordinal binary & High, Medium, Low & - & - & -0.800 & (0.55, 2.15) \\
    Density perception & Ordinal binary & High, Low & - & - & - & 1.55 \\

    \hline
    \end{tabular}
    \label{Table3}
\end{table}

As mentioned in the literature, it is common to evaluate the impact of all logical explanatory variables on the outcome, guided by prior knowledge. However, it is of vital importance to consider that conditioning on certain variables can inadvertently create non-causal paths between variables of interest. In addition, avoiding conditioning on non-collider variables keeps non-causal paths between the variables of interest unblocked. In other words, based on the main structures mentioned in Section \ref{IDSCM}, although conditioning on explanatory variables allows us to block non-causal paths that may exist between policy variables and outcomes, the selection of conditioning variables to avoid introducing spurious associations is crucial for causal analysis. In the simulated dataset, density perception acts as a collider variable in such a way that conditioning on that can create bias in the estimation of causal effects between stress level and wait time. 

\subsection{Comparison with non-causal based model}
We recover the direct effect of stress level on wait time using two distinct approaches: 1) modelling the association using the aforementioned typical variable selection method, and 2) modelling the causal mechanism based on the causal graph. In both approaches, we utilize ResNet-based frameworks, albeit with different model specifications. More precisely, in the first approach, all reasonable variables are included in Ordinal-Reslogit. In contrast, in the second approach, the model is specified based on the causal graph and back-door criteria. {It is worth noting that, because a residual neural network component is incorporated in both model structures, any differences in performance between the two models can be primarily attributed to variable selection method and observed confounding.} The mean value of the estimated parameter for both approaches is provided in Table \ref{result-simulatedresult}. 

\begin{table}[!htbp]
    \centering
    \caption{Comparison between causal and and non-causal-based models}\label{result-simulatedresult} 
    \scriptsize
    \begin{adjustbox}{width=0.75\textwidth}
    \begin{tabular}{ccccc}
    \hline
    \addlinespace
    \textbf{Cause} & \textbf{Effect} & \textbf{True causal effect} & \textbf{Ordinal-ResLogit} & \textbf{{{Causal model}}} \\
    &&&& \textbf{{{in \emph{CAROLINA}}}}\\
    %\addlinespace
    \hline
    
    Stress level & Wait time & 0.600 & 0.153 & 0.410 \\

    \hline
    \end{tabular}
    \end{adjustbox}
    \label{Table4}
\end{table}

Comparing the result of the two approaches with the true effect demonstrates that the typical approach in which we include the density perception (collider variable) into utility function, leads to a much higher error in estimation, approximately 42.9\%. However, \emph{CAROLINA} which is built upon the causal structure and does not consider the effect of the collider, achieves more accurate and unbiased results.

\subsection{Comparison with other causal based models}
{We compare the performance of \emph{CAROLINA} framework with other deep learning based causal models recently proposed in transportation literature, including STNSCM} \parencite{yang2023counterfactual}{, CGT} \parencite{deng2023spatio}{, and  MSCT} \parencite{li2024msct}. {Unlike \emph{CAROLINA}, these models were not designed to be interpretable and originally dealt with time-series continuous data (e.g., traffic flow). In order to be able to compare with CAROLINA, we modified their structures to be able to model the discrete behavioural choices. Table} \ref{tab:predict} {shows that the causal model in \emph{CAROLINA} framework is able to match the train and test accuracy of other algorithms with an 80-20 split and produces better results for the Binary Cross-Entropy (BCE) loss associated with the prediction of density perception in Figure }\ref{DAG-simulated}{. We also note that the difference between train and test accuracy is minimal.}

% Implementation details
%Deng: inferring latent confounders via an auxiliary neural network module and feeding both the treatment ($X$) and the estimated confounder ($Z$) into a Neural Structural Equation.
%Yang: introducing a learnable attention/masking gate before the predictor. The model learns to ``mask'' the input feature to find a robust representation.
%Li: adapting this by forcing the deep representation through a ``treatment-aware'' balancing layer. It re-weights the internal hidden states to simulate the MSM balancing property.

\begin{table}[H]
    \centering
    \caption{Accuracy comparison with other causal-based models.}
    \label{tab:predict}
    \scriptsize
    \renewcommand{\arraystretch}{1.2}
    \begin{tabular}{lcccc}
        \hline
         & \textbf{STNSCM} & \textbf{CGT} & \textbf{MSCT} & \textbf{Causal model}\\ 
         &&&&\textbf{in} \emph{\textbf{CAROLINA}}\\ \hline
        Train & 0.754 & \textbf{0.756} & 0.754 & 0.754\\
        Test & 0.745 & 0.740 & \textbf{0.747} & 0.746 \\
        BCE Loss & 0.520 & 0.525 & 0.519 & \textbf{0.513} \\ \hline
    \end{tabular}
\end{table}

{Furthermore, to compare the robustness of the complete \emph{CAROLINA} framework with other models, we used four different tests from recent literature, including a) Out-Of-Distribution (OOD) generalization} \parencite{krueger2021out}{, b) Brier score} \parencite{naumann2026wasserstein}{, c) Noise/Confusion sensitivity} \parencite{keogh2024prediction}{, and d) Wasserstein Individual Treatment Effect (ITE) distance} \parencite{rathod2026sample}{. OOD generalization measures the effect of extreme policy shocks, especially those not seen in the training data.} %For this metric, the variance in the counterfactual MSE at increasingly extreme $do(X)$ bounds is calculated. 
{The Brier score measures the difference between predicted counterfactual probabilities and actual choices. Noise/confusion sensitivity measures how the ITE prediction degrades when additional unobserved noise or bias is introduced into the system (e.g., simulating missing key variables). Wasserstein ITE measures the ability to captures the heterogeneity of the population by computing the 1D Wasserstein distance between predicted and true ITE distribution. Table} \ref{tab:robust} {shows that the \emph{CAROLINA} framework outforms benchmark models by showing higher consistency across various extreme policy shocks, representing the probability distribution of the choices better, showing a lower sensitivity to a random Gaussian noise injected into the input layer during inference, and capturing the heterogeneity of the population better.} 

\begin{table}[H]
\centering
\scriptsize
\setlength{\tabcolsep}{8pt}
\renewcommand{\arraystretch}{1.2}
\caption{Robustness comparison with other causal-based models}
\label{tab:robust}
\begin{tabular}{lcccc}
\hline
 & \textbf{STNSCM} & \textbf{CGT} & \textbf{MSCT} & \emph{\textbf{CAROLINA}} \\
\hline
OOD generalization & 0.0005 & 0.0005 & 0.0004 & \textbf{0.0001} \\
Brier score & 0.193 & 0.193 & 0.192 & \textbf{0.160} \\
Noise/Confusion sensitivity & 0.085 & 0.105 & 0.088 & \textbf{0.040} \\
Wasserstein ITE distance & 0.148 & 0.148 & 0.145 &  \textbf{0.026} \\
\hline
\end{tabular}
\end{table}

\subsection{Recovering from erroneous DAG discovery}
{For the false positive error, we add a case where the causal discovery algorithm mistakenly adds a connection between a fake/spurious variable and density. To evaluate the performance of the four methods, we use two statistical tests, including a) Intervention Invariance Score (IIS)} \parencite{fan2026frequency}{, and b) Partial Regression Coefficient (PRC)} \parencite{abdi2004partial}{. IIS evaluates how stable a causal model remains when actively manipulated or intervened upon by a fake variable. The fake/spurious variable should have no effect under do(). PRC represents the change in the dependent variable for a unit change in a fake variable, while holding other predictors constant. The ideal value in both tests should be zero. As shown in Table} \ref{tab:false}{, the tests suggest that \emph{CAROLINA} framework performs better than all the other deep learning based causal methods.} 

\begin{table}[H]
\centering
\scriptsize
\setlength{\tabcolsep}{8pt}
\renewcommand{\arraystretch}{1.2}
\caption{Comparison of performance in recovering from false positive.}
\label{tab:false}
\begin{tabular}{lcccc}
\hline
 & \textbf{STNSCM} & \textbf{CGT} & \textbf{MSCT} & \emph{\textbf{CAROLINA}} \\
\hline
IIS & 0.0984 & 0.0936 & 0.0997 & \textbf{0.0935}  \\
PRC  & 0.0721 & 0.0721 & 0.0721 & \textbf{0.0715}  \\
\hline
\end{tabular}
\end{table}

{In the case of false negative error, the discovery algorithm may miss a legitimate edge (e.g., between ``number of cars'' and ``stress level'' in Figure} \ref{DAG-simulated} {). Based on the domain knowledge, the modeller should be able to identify such an error and fix it by manually adding the edge before moving to Step 2. It can also be handled by making the causal discovery algorithm to be highly sensitive by tuning the hyperparameter for high recall. This will result in an ``over-connected'' DAG, which can be handled by the parameter estimation step of \emph{CAROLINA}, as shown in the tests for false positive error.}

\section{Case Studies} \label{data}
In order to analyze the performance of our proposed models, we use two observational datasets: 1) a pedestrian crossing behaviour dataset collected based on Virtual Immersive Reality Experiment (VIRE) \parencite{farooq2018virtual} in a controlled environment, and 2) a revealed preference for travel behaviour in London \parencite{hillel2018recreating}. 

\subsection{Datasets}
The first dataset consists of 1,406 responses for pedestrian crossing behaviour under different scenarios involving interactions with current traffic conditions, fully automated vehicles, and mixed traffic conditions \parencite{mudassar2021analysis}. These scenarios were defined based on various controlled factors presented in Table \ref{pedestriandataset}. In addition, Galvanic Skin Response (GSR) sensors were utilized during the experiments to measure the relative stress levels of the participants. The GSR sensor uses a small electrical charge to measure the amount of sweat on an individual's finger, where a higher charge indicates greater sweat production. It is important to note that in our study, the mean stress level of pedestrians in each scenario was normalized based on their minimum and maximum observed stress levels within that scenario. Detailed information on the definitions of scenario can be found in \cite{mudassar2021analysis}.

Moreover, we use an openly available dataset, the London Travel Demand Survey (LTDS), to investigate travel demand in the metropolitan area of London \parencite{hillel2018recreating}. This dataset combines the individual trip records from April 2012 to March 2015 with corresponding trip trajectories and mode alternatives obtained from a directions Application Programming Interface (API). Our focus in this study is on non-mandatory trips, where individuals have flexibility in selecting their destination based on their preferences and trip characteristics. Table \ref{londondata} presents a list of explanatory variables used in this study.

It is worth mentioning that in this study, pedestrian wait time, stress level and travel distance variables are discretized into discrete categories using Jenks Natural Breaks classification method and logical facts \parencite{jiang2013head}. This method categorizes a given dataset into multiple categories, with the objective of minimizing the variance within each group and maximizing the variance between different groups. The optimum thresholds of different categories for these variables are presented in Table \ref{pedestriandataset} and \ref{londondata}.

\begin{table}[H]
    \caption{Description of explanatory variables of pedestrian crossing behaviour}\label{pedestriandataset}
	\begin{center}
           \begin{adjustbox}{width=0.8\textwidth}
		\begin{tabular}{l p{9.5cm} l >{\centering\arraybackslash}p{2cm} c l}
		    \\\hline
      \hline
			 \textbf{Variable} & \textbf{Description} & \textbf{Mean} & \textbf{Standard deviation} \\\hline
			\emph{\textbf{Street Characteristics}} \\
                lane Width & & & \\
			\multicolumn{1}{c}{\small{Low}}   & \small{1: If      the lane width is less than 2.75 meter, 0:otherwise} &          0.334  & 0.472\\
                \multicolumn{1}{c}{\small{Medium}} & \small{1: if the lane width equals to 2.75 meter, 0: otherwise} & 0.307 & 0.462\\
    		\multicolumn{1}{c}{\small{High}} & \small{1: If the lane width is greater than 2.75 meter, 0:otherwise} & 0.359 & 0.480\\
                Road Type & & & \\
			\multicolumn{1}{c}{\small{Two way with a median}} & \small{1: If the road type is two way with median, 0:otherwise} & 0.338 & 0.473 \\
			\multicolumn{1}{c}{\small{Two way}}   & \small{1: If the road type is two way, 0:otherwise} & 0.317 & 0.466 \\
			\multicolumn{1}{c}{\small{One way}} & \small{1: If the road type is one way, 0:otherwise} & 0.344 & 0.475\\
                \small{Traffic Density} & \small{Density of road (veh/hr/ln)}  & 19.835 & 7.158\\
   \hline
			\emph{\textbf{Traffic Condition}} \\
                Automated Condition & & & \\
			\multicolumn{1}{c}{\small{Mixed}}   & \small{1: If traffic in scenario consists of automated vehicles and human-driven vehicles, 0:otherwise} & 0.045 & 0.207 \\
			\multicolumn{1}{c}{\small{Fully automated }}  & \small{1: If traffic in scenario consists of only automated vehicles, 0:otherwise} & 0.925 & 0.263 \\
			\small{Human-driven condition}   & \small{1: If traffic in scenario consists of only human-driven vehicles, 0:otherwise} & 0.029 & 0.170\\ \hline
			\emph{\textbf{Socio-demographic}} \\
                \small{Age group} & & & \\
			\multicolumn{1}{c}{\small{ 18-25}} & \small{1: If participant’ age is between 18 and 25, 0:otherwise} & 0.257  & 0.437 \\
			\multicolumn{1}{c}{\small{ 25-29}} & \small{1: If participant’ age is between 25 and 29, 0:otherwise} & 0.285  & 0.451 \\
			\multicolumn{1}{c}{\small{ 30-39}} & \small{1: If participant’ age is between 30 and 39, 0:otherwise} & 0.339  & 0.473 \\
                \multicolumn{1}{c}{\small{ 40-49}} & \small{1: If participant’ age is between 40 and 49, 0:otherwise} & 0.034  & 0.182 \\
			\multicolumn{1}{c}{\small{ Over 50}} & \small{1: If participant’ age is more than 50, 0:otherwise} & 0.084  & 0.278 \\
			\small{Gender} & \small{1: If participant is female, 0:otherwise} & 0.382  & 0.486\\
			\small{Driving Licence} & \small{1: If participant has a driving license, 0:otherwise} & 0.895  & 0.307\\
                Car Access & & & \\
			\multicolumn{1}{c}{\small{No car}} & \small{1: If participant has no car in the household, 0:otherwise} & 0.302  & 0.345\\
			\multicolumn{1}{c}{\small{One car }}& \small{1: If participant has one car in the household, 0:otherwise} & 0.345  & 0.476\\
			\multicolumn{1}{c}{\small{Over one car}} & \small{1: If participant has more than one car in the household, 0:otherwise} & 0.353  & 0.478\\
                Travel Mode & & & \\
			\multicolumn{1}{c}{\small{Active mode}} & \small{1: If participant use active modes regularly, 0:otherwise} & 0.258  & 0.438\\
			\multicolumn{1}{c}{\small{Private car mode}} & \small{1:If participant use private car regularly, 0:otherwise} & 0.332  & 0.471\\
			\multicolumn{1}{c}{\small{Public mode}} & \small{1: If participant use transit regularly, 0:otherwise} & 0.409  & 0.492\\
			\hline
                \emph{\textbf{Crossing Attributes}} \\
                \small{Wait Time} & \small{High: If participant’s Wait Time on the sidewalk is greater than 5 seconds, Low: otherwise} & 0.357  & 0.479 \\
			\small{Stress Level} & \small{High: If participant’s normalized stress is greater than 0.59, Low: otherwise} & 0.252  & 0.435 \\
                \hline
			\emph{\textbf{Environment Condition }} \\
			\small{Night} & 1\small{: If the time of scenario is night, 0:otherwise} & 0.371  & 0.483\\
			\small{Snowy} & \small{1: If the weather of scenario is snowy, 0:otherwise} & 0.293  & 0.455\\
			\hline
		\end{tabular}
          \end{adjustbox}
	\end{center}
 \label{Table5}
\end{table}

\begin{table}[h]
    \caption{Description of explanatory variables of London travel behaviour dataset}\label{londondata} 
    \begin{center}
    \begin{adjustbox}{width=0.7\textwidth}
        \begin{tabular}{l p{9.5cm} l >{\centering\arraybackslash}p{2cm} c l}
            \\\hline
            \textbf{Variable} & \textbf{Description} & \textbf{Mean} & \textbf{Standard deviation} \\\hline
            \emph{\textbf{Travel Attributes}} \\
            
            Travel Mode & & & \\
            \multicolumn{1}{c}{\small{Private car}}   & \small{1: if individual uses private car, 0: otherwise} &   0.478 & 0.450\\
            \multicolumn{1}{c}{\small{Public transit}} & \small{1: if individual uses public transportation, 0: otherwise } & 0.296 & 0.456\\
            \multicolumn{1}{c}{\small{Active modes}} & \small{1: if individual uses bike or walking, 0: otherwise } & 0.226 & 0.456\\
            Travel Cost & & & \\
            \multicolumn{1}{c}{\small{Transit cost}} & \small{Travel cost for transit mode (£)} & 1.456 & 1.287 \\
            \multicolumn{1}{c}{\small{Driving cost}}   & \small{Total travel cost for private car mode (£)} & 1.423 & 2.853 \\
            Travel Distance & \small{Long: more than 7.8 kilometres away from their starting point, Medium: otherwise} & 0.529 & 0.499 \\
            \hline
            
            \emph{\textbf{Socio-demographic}} \\
            \small{Age Group} & & & \\
            \multicolumn{1}{c}{\small{ 18-30}} & \small{1: if individual’s age is between 18 and 30, 0: otherwise} &0.185 & 0.388 \\
            \multicolumn{1}{c}{\small{ 30-45}} & \small{1: if individual’s age is between 30 and 45, 0: otherwise} & 0.315  & 0.465 \\
            \multicolumn{1}{c}{\small{45-60}} & \small{1: if individual’s age is between 45 and 60, 0: otherwise} & 0.232  & 0.422 \\
            \multicolumn{1}{c}{\small{ Over 60}} & \small{1: if individual’s age is more than 60, 0: otherwise} & 0.268  & 0.443 \\
            \small{Gender} & \small{1: if individual is female, 0: otherwise} & 0.557  & 0.497\\
            \small{Disability} & \small{1: If participant has a disability transit fare, 0:otherwise} & 0.895  & 0.307\\
            \small{Driving License} & \small{1: if individual has a driving license, 0: otherwise} & 0.718  & 0.450\\
            \small{Unrestricted Car Access} & \small{1: if there is one or more than one car in the household, 0: otherwise} & 0.271  & 0.444\\
            \hline
        \end{tabular}
    \end{adjustbox}
    \end{center}
    \label{Table6}
\end{table}

\subsection{Model Hyperparameters}
%In the context of deep learning, the model hyperparameter setting plays an important role in determining the performance and generalization capabilities of the model. In the process of modelling, we divided the dataset into two subsets using a 70:30 training/validation split. In this study, 
We employ the random search technique to optimize the hyperparameters of our model. This involves defining a range of values for each hyperparameter and randomly selecting combinations from these ranges to train the model on the training dataset \parencite{bergstra2012random}. Table \ref{Hyperparameters} shows the range of hyperparameters used in the random search method. 

Note that in this study, we use the early-stopping approach to overcome the overfitting problem; however, other regularization techniques, that may further enhance the performance of learning models, are not used in our study. This decision was made to facilitate a fair comparison between our proposed model dealing with unobserved confounders through the deep component and the traditional behaviour modelling approach. We implement the structure of our proposed model using open-source deep learning libraries in Python. %The code for our experiments will be available on the lab's GitHub as the manuscript is published.

\begin{table}[h]
    \centering
    \caption{Range of model hyperparameters}\label{Hyperparameters}
    \begin{adjustbox}{width=0.3\textwidth}
    \begin{tabular}{ccc}
    \hline
    %\addlinespace
    \textbf{Parameters} & & \textbf{Value}  \\
    %\addlinespace
    \hline
    %\addlinespace
    Residual Layers & & \{2, 4, 8, 16, 32\} \\
    %\addlinespace
    %\addlinespace
    Learning rate	& & \{0.01, 0.001\} \\
    %\addlinespace
    %\addlinespace
    Batch size & & \{32, 64\} \\
    %\addlinespace
    %\addlinespace
    Number of hidden layers & & \{4, 6\} \\
    %\addlinespace
    %\addlinespace
    Threshold of ordinal categories & & [0.3, 0.6]\\
    
    \hline
    \end{tabular}
    \end{adjustbox}
    \label{Table7}
\end{table}

\section{Results and Discussion} \label{result}
This section is divided into three subsections in which the results of causal discovery, interpretable deep structural causal modelling, and the generative counterfactual model for each dataset are provided.

\subsection{Directed Acyclic Graphs (DAGs) Learning}
In order to learn Directed Acyclic Graphs (DAGs) for each dataset, we focus on the causal discovery algorithms that are based on sufficiency assumptions, meaning that the observed data contains sufficient information to determine the causal relationships between variables. In other words, in this step, we assume that the causal graphs are not influenced by any unobserved confounders. 
%PC and FGES described in Section \ref{Causal Discovery} are two popular causal discovery algorithms based on sufficiency assumptions; nevertheless, in the literature, there is no standard causal discovery method. 
In this study, we used the FGES algorithm for two reasons. The FGES enjoys parallelization techniques in optimization and searching the causal graph. In addition, although the PC algorithm and its variants consist of two phases, searching a skeleton and orienting the edges, they are not appropriate for big datasets as the second phase may be removed, resulting in an undirected graph \parencite{nogueira2022methods}. Notably, in DAG learning, we incorporate asymmetrical analysis within the FGES algorithm to determine the true directional relationship between two variables based on the model score. Moreover, we incorporate expert knowledge into the FGES algorithm to decrease the challenges associated with the Markov Equivalence Class problems.

\subsubsection{Pedestrian crossing behaviour}
To learn the causal structures of pedestrian behaviour in the presence of automated vehicles, the following prior knowledge is considered:
\begin{itemize}
    \item The street characteristics, traffic condition, and environment condition variables are only allowed to cause changes in pedestrian wait time and stress level. 
    \item The meaningless relationship between some socio-demographic variables like age $\rightarrow$ gender or driving licence $\rightarrow$ age are prohibited. However, the logical cause and effect relationships such as age $\rightarrow$ driving licence are allowed.
\end{itemize}

\begin{figure}[!ht]
  \begin{center}
   \includegraphics[width=0.6\textwidth]{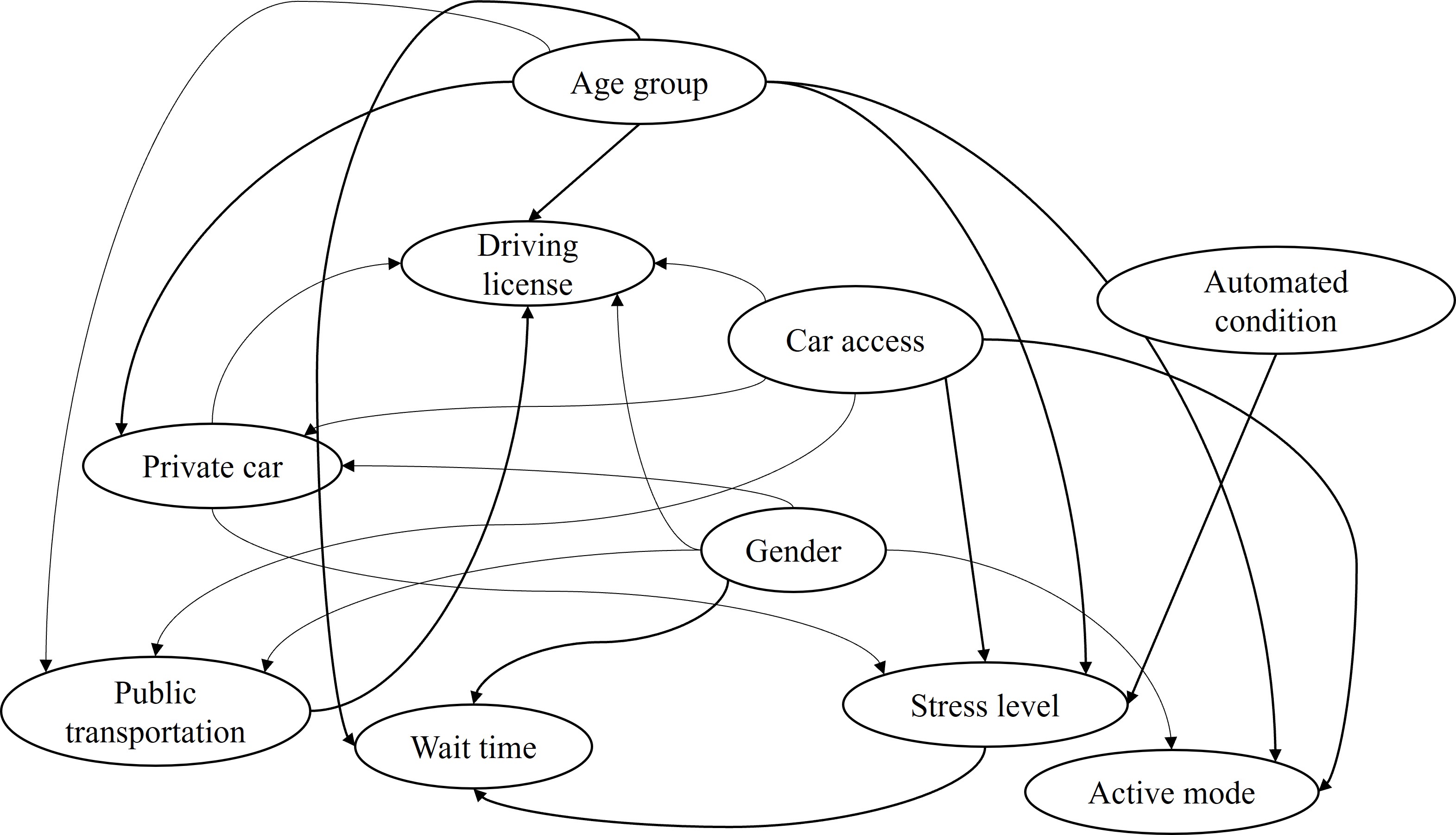}
   \caption{Causal structure of pedestrian crossing behaviour}\label{DAGFGES-ped}
   \end{center}
\end{figure}

Figure \ref{DAGFGES-ped} illustrates the causal structure obtained from the FGES algorithm for the pedestrian crossing behaviour. Interestingly, this causal structure confirms our assumption in previous studies stating that pedestrian stress level influences their waiting time \parencite{kamal2022debiased}. Furthermore, a first glance at this graph illuminates the need for causality analysis. In fact, in most travel behaviour analyses in transportation, we assume all observed variables as pre-treatment of outcome in policy assessment and may wrongly interpret associations as the likely reasons for changes in the outcome. However, this assumption would not be representative of the true behaviour-generating process. For instance, in the causal structures of pedestrian behaviour, there is no direct relationship between the number of cars in the household and pedestrian wait time. However, the graph shows the number of cars directly influences pedestrian stress level and then the effect of this variable is transferred to pedestrian wait time through their stress level. Conversely, in typical behavioural modelling in transportation, by conditioning on both the number of cars and stress level variables, the indirect effect of the number of cars on pedestrian waiting time is blocked and cannot be estimated.

In this causal graph, age, gender, number of cars and traffic condition are only exogenous variables and other variables play roles as both exogenous $(x_{ex})$ and endogenous $(x_{en})$ variables in our framework. In this study, our goal is to examine how variations in pedestrian stress levels (policy) influence waiting time (outcome), represented as (stress level $\rightarrow$ waiting time). As shown in Figure \ref{DAGFGES-ped}, between these variables, the fork structure can be observed due to peoples' age. In other words, this variable is the common cause of generating waiting time and stress level, potentially confounding the true impact of variables on each other. These findings underscore the vital importance of identifying causal structures in behavioural modelling, particularly in the study of policy assessment.

\subsubsection{London travel behaviour}
\begin{figure}[h]
  \begin{center}
   \includegraphics[width=0.65\textwidth]{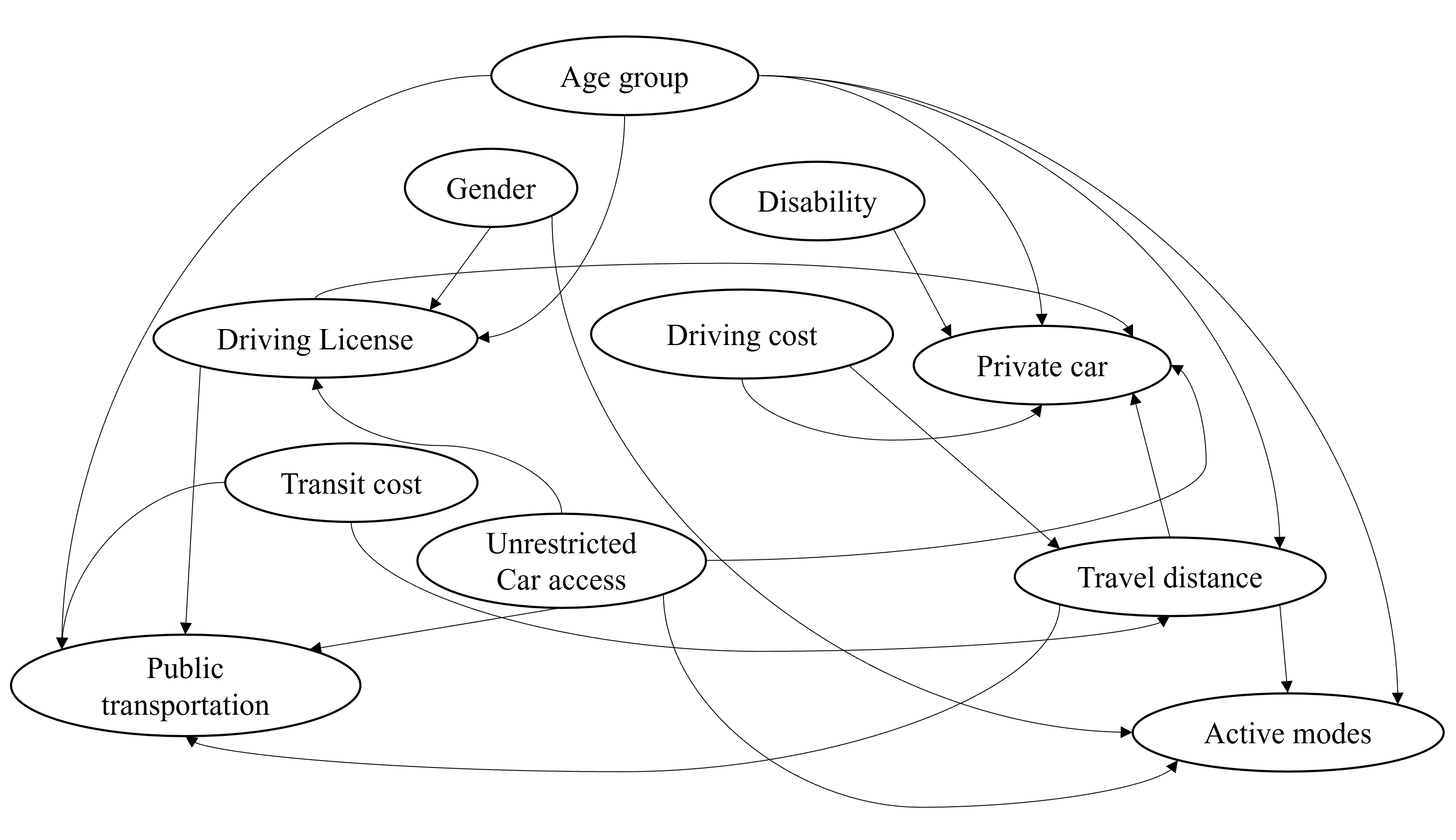}
   \caption{Causal structure of London travel behaviour }\label{DAGFGES-London}
   \end{center}
\end{figure}

Figure \ref{DAGFGES-London} demonstrates the behaviour-generating process in the London travel dataset obtained from the FGES algorithm. For this dataset, we also contemplate the following prior knowledge:

\begin{itemize}
    \item Regarding some socio-demographic variables such as age, gender, disability and car access variables are treated to be exogenous variables $(x_{ex})$ in the model. In other words, these variables are not affected by any other variable (parents) in the causal structures.
    \item The meaningless relationships between some socio-demographic variables like age $\rightarrow$ gender are not allowed.
    \item Any direct connection between travel modes, including private car, public transit, and active modes, is prohibited. Note that in the modelling the cross-effect of mode choices is captured by using deep residual layers. 
    \item The connection between an alternative and non-alternative-specific attributes (attributes of other travel modes) is restricted.
    \item The cost of both driving and transit are not allowed to be influenced by any other variables in this dataset. 
\end{itemize}

The results uncover that for non-mandatory trips, the arbitrary choice of travel distance is a cause of the choice of travel modes. However, other common causes between these two variables may also connect them through a non-causal path. This result does not mean that our study criticizes the previous studies in which researchers evaluated the effect of travel mode choice on travel distance. Such variations might arise due to two reasons. First, each causal structure describes one population, implying that causal graphs of similar problems should be localized for each population \parencite{pearl2009causality}. Secondly, unlike previous studies that emphasized association-based analysis, we prioritize causality analysis for policy assessment. The non-causal studies are able to reveal associations between variables, however, these associations must not be interpreted as causal associations or used for policy assessment. In the next section, we estimate the causal direction between travel distance (policy) and three travel modes (outcome) after accounting for confounders.   

\subsection{Cause and Effect Analysis}
The causal graphs obtained from the discovery process are fed into the deep structural causal model of \emph{CAROLINA} in order to account for both observed and unobserved confounders. To emphasize the impact of unobserved confounders, we compare \emph{CAROLINA} with a Structural Causal Model that relies on a predetermined utility function without incorporating deep learning components. This model is built on Multinominal Logit (MNL) and Ordered Logit Models that are widely used in the literature. This model is called as a SCM-Logit model in this study. Our preliminary assumption is that both \emph{CAROLINA} and SCM-Logit specified based on the same causal structures or DAGs, ensuring that both models handle observed confounders in the same way. However, the key distinction is that \emph{CAROLINA} incorporates unobserved confounders, while SCM-Logit does not.

In the study of pedestrian crossing behaviour, in order to estimate the causal effect of stress levels on pedestrian wait times, our model approach takes into account the causal relationship between travel mode choice, pedestrian stress level, waiting time and other exogenous variables. In fact, based on the causal structure, for instance, an individual's inclination towards private car usage may impact their stress levels on a sidewalk. Concerning the second dataset, we analyze the causal process in which individuals choose medium and long travel distances and examine how these choices, along with their possession of a driving license, directly influence their travel model choice. Table \Ref{DSEM-ped} and \Ref{DSEM-London} present the results obtained for pedestrian crossing and travel behaviour datasets, respectively.

Due to higher total number of parameters associated with the deep layers, for the model selection and analysis of the goodness of fit, we used the Akaike Information Criterion (AIC) values (Equation \ref{eqAIC}).

\begin{equation}
    AIC = -2LL(\beta,\theta)  + 2 B
    \label{eqAIC}
\end{equation}

Note that previous hybrid behavioural models have also used the Bayesian Information Criterion (BIC) by not considering the weights of the hidden layers of the neural network, but only considering the parameters directly associated with the independent variables. However, we are of the view that AIC is a better statistic to use where neural networks and logit models are applied and compared on large datasets, while representing all the parameters of the models. The lower AIC value of \emph{CAROLINA} in Table \ref{Table1} shows that \emph{CAROLINA} outperforms the SCM-Logit. In addition, concerning the accuracy, we use Mean Prediction Error (MPE) on the validation dataset:

\begin{equation}
    MPE = \displaystyle \frac{1}{N} \sum_{n=1}^N \mathbb{I}(y_{n} \neq {\tilde{y}}_{n})
\end{equation}

where $y_{n}$ and $\tilde{y}_{n}$ are the actual and predicted choices for each individual $n$ and indicator index $\mathbb{I}$ equals 1 when the expression inside is true and otherwise, it is 0. In terms of pedestrian crossing behaviour, we observe approximately 19.4\% and 33.4\% and 3.8\% differences in the accuracy of discrete travel mode choice, stress level, and waiting time causal mechanisms, respectively. Furthermore, in the travel behaviour data, a notable improvement in the accuracy of model is observed by 27.2\% in the prediction of travel distance. Table \Ref{DSEM-ped} and \Ref{DSEM-London} show significant exogenous variables in causal mechanisms with their t-statistic information. In our proposed model, based on back-door criteria, each causal mechanism only takes variables identified as their parent within the causal structure. In addition, some variables were retained in the model to control for confounding effects in each causal mechanism, even if their associated parameters were statistically not significant. In fact, although in economics, association-based analyses often discard such variables, leading to short model specification, causal analysis takes a different approach and the goal is to block confounding associations, even if they are minimal.

Concerning pedestrian crossing behaviour, our results reveal that daily using private cars positively influences pedestrian stress levels. In fact, among other alternatives, the habitual use of private cars significantly influences the level of stress experienced by individuals on the sidewalk. {In addition, the CAROLINA results indicate that higher accessibility to a private car increases the probability of a higher stress level. A consistent result was also observed in the other model, where this variable affected stress level with a similar direction \parencite{kamal2024ordinal}.} Another noteworthy result of the pedestrian crossing study shows that their stress level significantly changes their decisions regarding the start time of crossing. In other words, a high level of stress, which can be influenced by pedestrians' age group, gender, travel pattern, and traffic conditions, results in longer waiting time on the sidewalk. However, the reverse is obtained in the SCM-Logit. This result underscores the impact of confounders, as the SCM-Logit model accounts for observed confounders but fails to control for unobserved confounders, leaving unobserved non-causal associations unblocked. Our recent study uncovered a negative unobserved confounding correlation between waiting time and stress level using the same dataset \parencite{kamalFarooq_2025}. This association plays a significant role in confounding the direct effect of stress level on waiting time in the SCM-Logit model. In other words, the differences in results between SCM-Logit and \emph{CAROLINA} stem from the deep learning component of causal mechanisms.  In \emph{CAROLINA}, this component blocks unobserved confounding effects, measured in \cite{kamalFarooq_2025}, whereas in the SCM-Logit model, these confounding effects remain unaddressed. In addition, the SCM-Logit result also contradicts our real-world expectations. In fact, typically, individuals experiencing stress are expected to hesitate to cross the street immediately and wait longer on the sidewalk. It is worth noting that regarding road-related attributes our causal discovery approach only reveals relationships between road-related attributes and the stress level. Based on our investigation, the same result is also obtained by using the PC algorithm. However, it is important to acknowledge that due to the limitations of the virtual reality dataset, further research is necessary to validate and expand upon any direction between other road-related attributes and stress levels or the duration of waiting time on the sidewalk.

\begin{table}[H]
\centering
\caption{Results of estimated models for pedestrian crossing dataset}\label{DSEM-ped}
\begin{adjustbox}{width=0.7\textwidth}
\begin{tabular}{lcccc}
  \hline
  \hline 
  \addlinespace
    % Table content goes here
  \textbf{Exogenous}  & \multicolumn{4}{c}{\textbf{Endogenous}}    \\
  \cline{2-5}
  \addlinespace[1.5ex]
  & \multicolumn{2}{c}{\textbf{\emph{{Causal model in CAROLINA}}}}  & \multicolumn{2}{c}{\textbf{SCM-Logit}} \\
  \hline
    & \multicolumn{4}{c}{\textbf{Travel Mode Choice}}  \\ 
    \addlinespace[1ex]
    & \emph{Public Transit} & \emph{Active} & \emph{Public Transit} & \textit{Active}\\
    \hline 
     \addlinespace[1ex]
    Age 25-29   &  0.950 (6.530$^{*}$)&	0.218 (0.985)	&	-0.380 (-2.152)	&-0.242 (-1.008) \\
    Age 30-39 &	-0.177 (-1.724)	& 0.054(0.360)	&	0.018 (0.154)	& -0.094 (-0.532)\\  
    Age over 50 & -0.419 (-1.869)&	-0.812 (-2.290)	&	-0.279 (-0.936)	& -0.168 (-0.438) \\
    Gender & 1.227 (6.750)	&1.125 (4.754)	&	0.919 (7.045)	&0.316 (1.590) \\
    One car & 0.052( 0.4902)	&-0.548 (-3.549)	&	-0.905 (-6.850)	& -0.464 (-2.520) \\
    Over one car & -0.776 (-7.366) &	-0.560 (-3.571)	&	-0.805 (-5.976) &	-0.431(-2.328) \\
    \hline
    & \multicolumn{4}{c}{\textbf{Stress Level}}  \\ 
    \hline
    Age 25-29 & \multicolumn{2}{c}{0.433(0.852)}  & \multicolumn{2}{c}{ 0.235 (1.429)} \\
    Age 30-39  & \multicolumn{2}{c}{0.991 (2.579)}  & \multicolumn{2}{c}{-0.491 (-3.832)} \\
    Age over 40-49  & \multicolumn{2}{c}{-}  & \multicolumn{2}{c}{0.190 (0.313)} \\
    Age over 50  & \multicolumn{2}{c}{0.949 (1.005)}  & \multicolumn{2}{c}{-} \\
    One car & \multicolumn{2}{c}{0.356 (1.697)}  & \multicolumn{2}{c}{0.268 (1.985)} \\
    Over one car & \multicolumn{2}{c}{0.395 (1.789)}  & \multicolumn{2}{c}{-0.070 (-0.502)} \\
    Private car & \multicolumn{2}{c}{0.083 (2.859)}  & \multicolumn{2}{c}{-1.354 (-8.496)} \\
    Fully automated condition & \multicolumn{2}{c}{0.539 (2.346)}  & \multicolumn{2}{c}{0.307 (3.693)} \\
    Mixed Traffic condition & \multicolumn{2}{c}{-0.037 (-0.916)}  & \multicolumn{2}{c}{0.025 (0.163)} \\
    \hline
    & \multicolumn{4}{c}{\textbf{Wait time}}  \\ 
    \hline
    Age 25-29  & \multicolumn{2}{c}{0.077(0.214)}  & \multicolumn{2}{c}{0.037 (0.279)} \\
    Age 30-39  & \multicolumn{2}{c}{1.056 (5.406)}  & \multicolumn{2}{c}{-0.088 (-1.860)} \\
    Age 40-49  & \multicolumn{2}{c}{-}  & \multicolumn{2}{c}{0.189 (0.388)} \\
    Age over 50  & \multicolumn{2}{c}{0.061(0.820)}  & \multicolumn{2}{c}{-0.284 (-1.166) } \\
    Gender & \multicolumn{2}{c}{0.770 (5.862)}  & \multicolumn{2}{c}{0.205 (1.836)} \\
    Stress level & \multicolumn{2}{c}{0.843 (5.154)}  & \multicolumn{2}{c}{-0.111 (-0.878)} \\
    \hline
    \hline
    No. observation & \multicolumn{2}{c}{1406}  & \multicolumn{2}{c}{1406} \\
    No. parameters  & \multicolumn{2}{c}{300}  & \multicolumn{2}{c}{16} \\
    Log-likelihood & \multicolumn{2}{c}{-1342.01}  & \multicolumn{2}{c}{-1833.13} \\
    AIC  & \multicolumn{2}{c}{3284.02}  & \multicolumn{2}{c}{3698.26} \\
    Mean Prediction Error (MPE) \\
    \multicolumn{1}{c}{\textbf{Travel Mode}} & \multicolumn{2}{c}{29.94\%}  & \multicolumn{2}{c}{49.36\%} \\
    \multicolumn{1}{c}{\textbf{Stress Level}} & \multicolumn{2}{c}{21.34\%}  & \multicolumn{2}{c}{54.78\%} \\
    \multicolumn{1}{c}{\textbf{Wait time}} & \multicolumn{2}{c}{41.08\%}  & \multicolumn{2}{c}{44.90\%} \\
  \hline
\end{tabular}
\end{adjustbox}
\label{Table1}
\begin{tablenotes}
\item \hspace{2cm} \tiny t-stats are outlined in the parenthesis.
\end{tablenotes}
\end{table}

Regarding the travel behaviour dataset, our findings indicate that increased household accessibility to private cars correlates with a higher probability of possessing a driving license. Subsequently, individuals with driving licenses demonstrate a stronger preference for both private cars and public transportation, although the inclination towards private cars is significantly higher. However, according to SCM-logit analysis, having a driving license likely leads to a greater inclination towards active modes of transportation compared to public transportation. Furthermore, the DAG indicates a direct impact between travel distance and people's preference towards travel modes. It is worth mentioning that the opposite direction may also exist, but the confounding variables serve as the primary reason for this direction, rather than a direct effect. In other words, based on our study, the reverse relationship reflects an overall association rather than a causal effect. 

Our result demonstrates that choosing more distant destinations for non-mandatory trips negatively changes the probability of opting for public transportation and increases people's tendency towards using private cars. In other words, in the travel behaviour-generating process, as the distance of the trip increases, there is a corresponding decrease in the individual's propensity towards modes of transportation that are environmentally friendly and promote sustainability. However, based on our study, we are not allowed to mistakenly extend this result in the reverse direction and conclude that choosing a private car causes travel to a farther destination. This is a key distinction between our analysis and typical travel behavioural analysis.

%\emph{CAROLINA} can be as complete as the traditional econometric models in terms of calculating the elasticity, substitution patterns and other economic information. Despite utilizing deep layers, our framework maintains the deterministic part of utility functions, allowing us to extract economic information. Furthermore, our framework enables us to confine the utility function of each alternative solely to its attributes, ensuring alternative-specific attributes assumption in the model specification. 

\begin{table}[H]
\centering
\caption{Results of estimated models for travel behaviour dataset}\label{DSEM-London}
\begin{adjustbox}{width=0.7\textwidth}
\begin{tabular}{lcccc}
  \hline
  \hline 
  \addlinespace
    % Table content goes here
  \textbf{Exogenous}  & \multicolumn{4}{c}{\textbf{Endogenous}}    \\
  \cline{2-5}
  \addlinespace[1.5ex]
  & \multicolumn{2}{c}{\textbf{\emph{{Causal model in CAROLINA}}}}  & \multicolumn{2}{c}{\textbf{SCM-Logit}} \\
  \hline
        & \multicolumn{4}{c}{\textbf{Driving Licence}}  \\ 
    \addlinespace[1ex]
    \hline 
     \addlinespace[1ex]
    Female & \multicolumn{2}{c}{-1.006 (-2.339)}  & \multicolumn{2}{c}{-0.207 (-3.550)} \\
    Age 30-45 & \multicolumn{2}{c}{1.325 (2.321)}  & \multicolumn{2}{c}{0.287 (3.776)} \\
    Age 45-60 & \multicolumn{2}{c}{1.455 (1.941)}  & \multicolumn{2}{c}{0.302 (2.459)} \\
    Age over 60 & \multicolumn{2}{c}{ 0.755 (1.055)}  & \multicolumn{2}{c}{0.153 (6.956)} \\
    Unrestricted Car Access & \multicolumn{2}{c}{5.839 (1.915)}  & \multicolumn{2}{c}{0.568 (4.501)} \\
    \hline
    & \multicolumn{4}{c}{\textbf{Travel Distance}}  \\ 
    \addlinespace[1ex]
    \hline 
     \addlinespace[1ex]
    Age 30-45 & \multicolumn{2}{c}{-0.015 (-4.442)}  & \multicolumn{2}{c}{-0.072 (-3.385)} \\
    Age 45-60 & \multicolumn{2}{c}{-0.045 (-11.461)}  & \multicolumn{2}{c}{-0.080 (-3.243)} \\
    Age over 60 & \multicolumn{2}{c}{-0.157 (-5.827)}  & \multicolumn{2}{c}{0.227 (10.239)} \\
    Transit cost & \multicolumn{2}{c}{-0.011 (-8.197)}  & \multicolumn{2}{c}{0.218 (6.343)} \\
    Driving cost & \multicolumn{2}{c}{1.664 (6.784)}  & \multicolumn{2}{c}{0.058 (16.196)} \\
    \hline
      & \multicolumn{4}{c}{\textbf{Travel Mode Choice}}   \\ 
    \addlinespace[1ex]
        & \emph{Public Transit} & \emph{Private Car} & \emph{Public Transit} & \textit{Private Car}\\
    \hline
    
    Age 30-45 &  -0.070 (-3.971)&	0.346 (1.371)	&	-0.114 (-5.002)	& -0.146 (-6.976) \\
    Age 45-60 &  -0.196 (-8.460)&	0.370 (1.046)	&	-0.116 (-4.173)	&-0.076 (-3.152) \\
    Age over 60 &  -0.828 (-2.282)&	-0.217 (-12.059)	&	0.129 (5.081)	& -0.064 (-2.837)\\
    Disability &  - &	-0.647 (-10.731)	&	-	& -0.245 (-4.705)\\
    Driving Licence &  0.104 (7.332)&	1.775 (5.329)	& -0.302 (-2.786)	& 0.262 (9.497)\\
    Driving cost &  - &	 -0.438 (-2.681)	& -	& -0.060 (-4.651)\\
    Transit cost &  -0.156 (-1.159)&	-	&	0.013 (2.0553)	& -\\
    Unrestricted Car Access &  -0.365 (-3.304)&	0.942 (1.458)	&	-0.092 (-3.498)	& 0.277 (12.858)\\
    Long Travel Distance &  -1.442 (-4.051)&	0.433 (1.064)	&	0.685 (4.148)	& 0.522 (3.232) \\
    \hline

    No. observation & \multicolumn{2}{c}{45,547 }  & \multicolumn{2}{c}{45,547} \\
    No. parameters  & \multicolumn{2}{c}{299}  & \multicolumn{2}{c}{26} \\
    Log-likelihood & \multicolumn{2}{c}{-47,055.94}  & \multicolumn{2}{c}{-70,701.37} \\
    AIC  & \multicolumn{2}{c}{94,709.88}  & \multicolumn{2}{c}{141,454.74} \\
    Mean Prediction Error (MPE) \\
    \multicolumn{1}{c}{\textbf{Driving Licence}} & \multicolumn{2}{c}{26.94\%}  & \multicolumn{2}{c}{27.87\%} \\
    \multicolumn{1}{c}{\textbf{Travel Distance}} & \multicolumn{2}{c}{10.87\%}  & \multicolumn{2}{c}{38.10\%} \\
    \multicolumn{1}{c}{\textbf{Travel Modes}} & \multicolumn{2}{c}{36.96\%}  & \multicolumn{2}{c}{39.54\%} \\
  \hline
\end{tabular}
\end{adjustbox}
\label{Table2}
\begin{tablenotes}
\item \hspace{2 cm} \tiny - t-stats are outlined in the parentheses.
\item \hspace{2 cm} \tiny - Marginal Rate of Substitution analysis based on this model can be found in Appendix \ref{app:econ}.
\end{tablenotes}
\end{table}

\subsection{Counterfactual Analysis}
\begin{figure}[h]
    \centering
    \subfloat{{\includegraphics[scale=0.35]{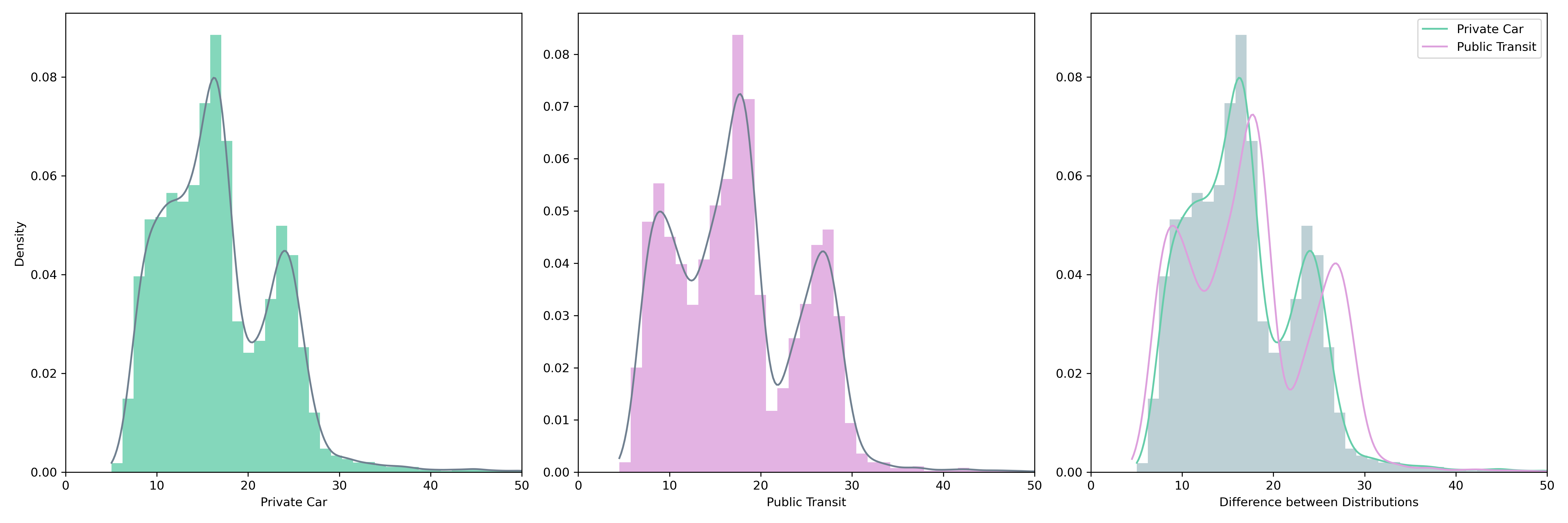}}}\\
    \subfloat{{\includegraphics[scale=.42]{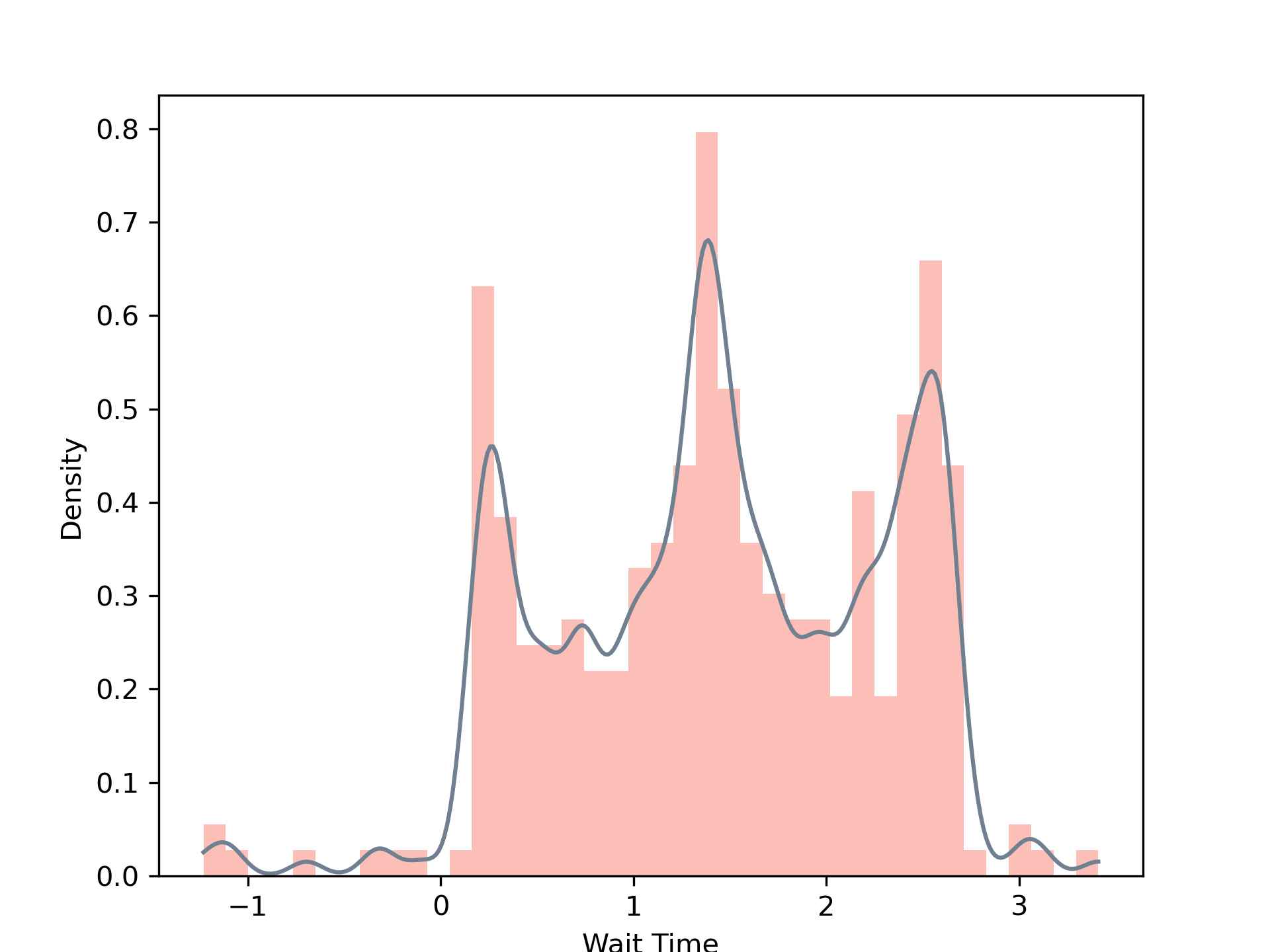}}}
    \caption{Distribution of the latent variable}
    \label{flowoutputdist}
\end{figure}

Here, we apply the generative counterfactual model structure to generate the counterfactual data for pedestrian and London mode choice datasets and compare factual and counterfactual behaviours. This model structure is able to address the fundamental challenge in causal inference and allows us to predict the counterfactual outcomes for each observation. We investigate the impact of specific interventions on pedestrian crossing behaviour and travel behaviour in London by: 1) exploring how an individual's wait time would change in a counterfactual scenario where stress levels were reduced to a low level for all individuals,
%sample by making the walking infrastructure safer 
and 2) assessing how individuals’ travel mode preferences would shift if recreational amenities, like shopping malls and restaurants, were accessible within a 7.8 kilometres ($\approx$5 miles) radius from their origin, representing medium distance trips. Based on our methodology, in both datasets, we first learn the distribution of the latent variables ($\varepsilon_i$) associated with the causal mechanisms of the aforementioned intervention distribution. The utilization of normalizing flow in our proposed model allows us to transition from a simple normal distribution assumption to a more representative distribution that captures the characteristics of latent behaviours in the dataset \parencite{kim2023finite}. Figure \ref{flowoutputdist} displays the visual representation of the distribution of the latent variable, which is captured by using the normalizing flow layers. The multimodal distributions illustrate that our datasets consist of multiple latent behaviours. In other words, there are multiple sub-population having different latent behaviour constructs. This property of the counterfactual model enables us to better capture underlying uncertainty and enhance the prediction of individual-level counterfactual behaviour. Furthermore, comparing the distributions of latent terms in the utility functions of private cars and public transit reveals differences. This result shows that these latent terms likely encompass the influence of mode-specific attributes.

Based on the second step of the counterfactual model, the structural causal mechanisms are changed based on the desired manipulation. This step is identical to the policy implementation in such a way that the stress level is set to a low level and travel distance is reduced to medium for all individuals in the population sample of dataset, respectively. Eventually, we utilize the deep structural causal model to predict the counterfactual outcome for each individual in the manipulated system.

\begin{figure}[!h]
    \centering
    {{\includegraphics[scale=.375]{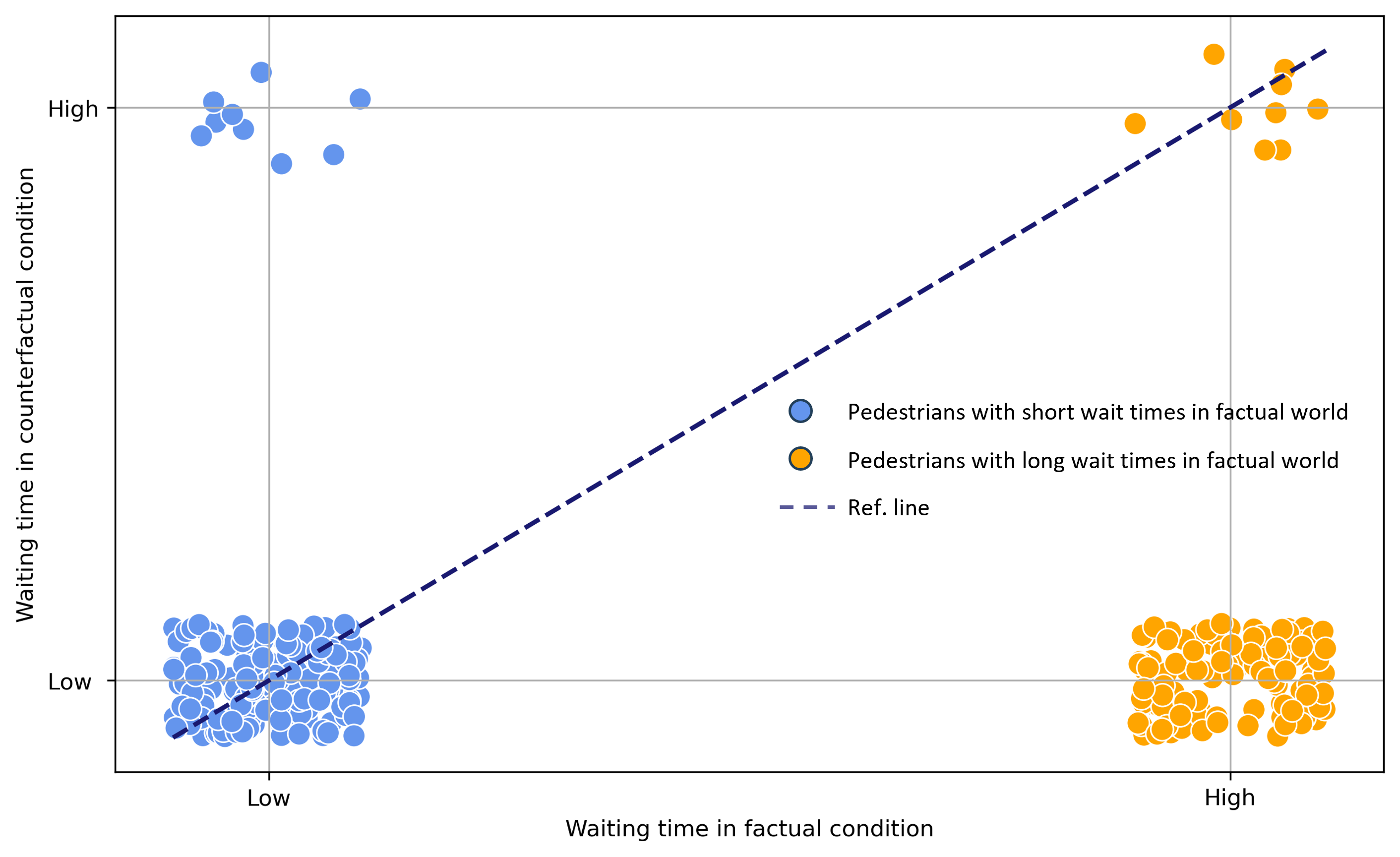}}}
    {{\includegraphics[scale=.4]{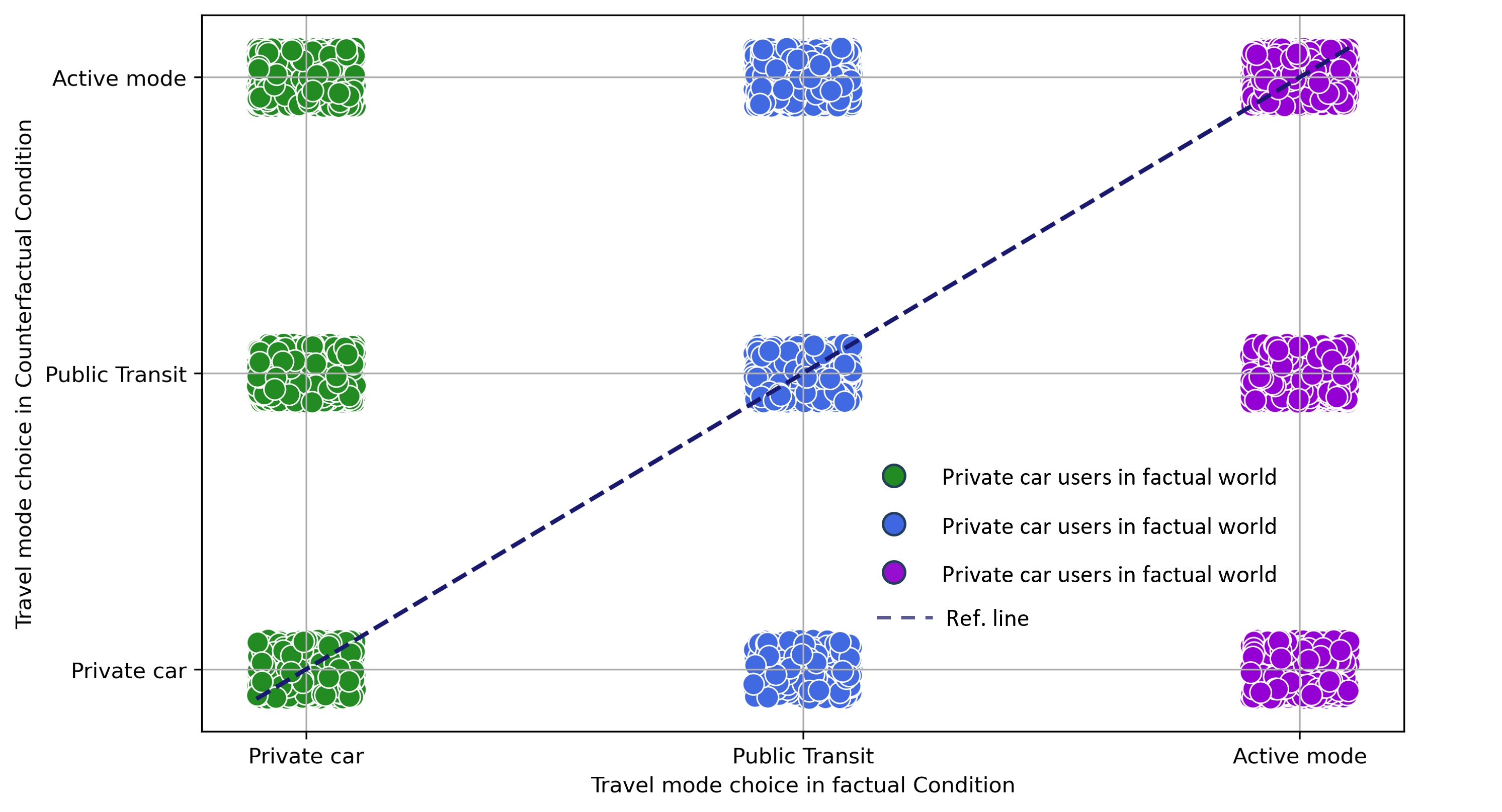}}}
    \caption{Individual treatment effect}
    \label{Individualcountefactualchart}
\end{figure}

\begin{figure}[!h]
    \centering
    {{\includegraphics[scale=.55]{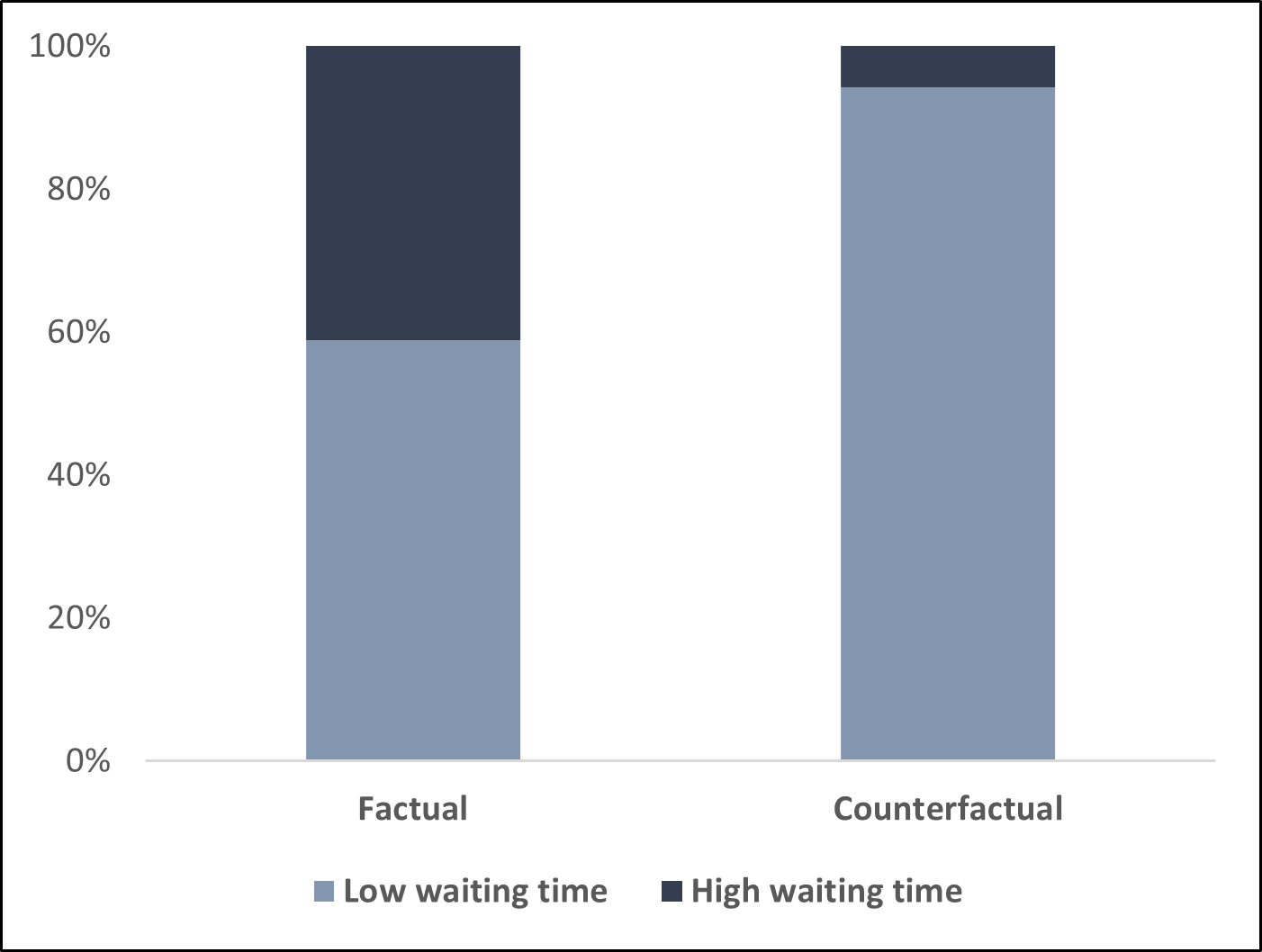}}}
    {{\includegraphics[scale=.55]{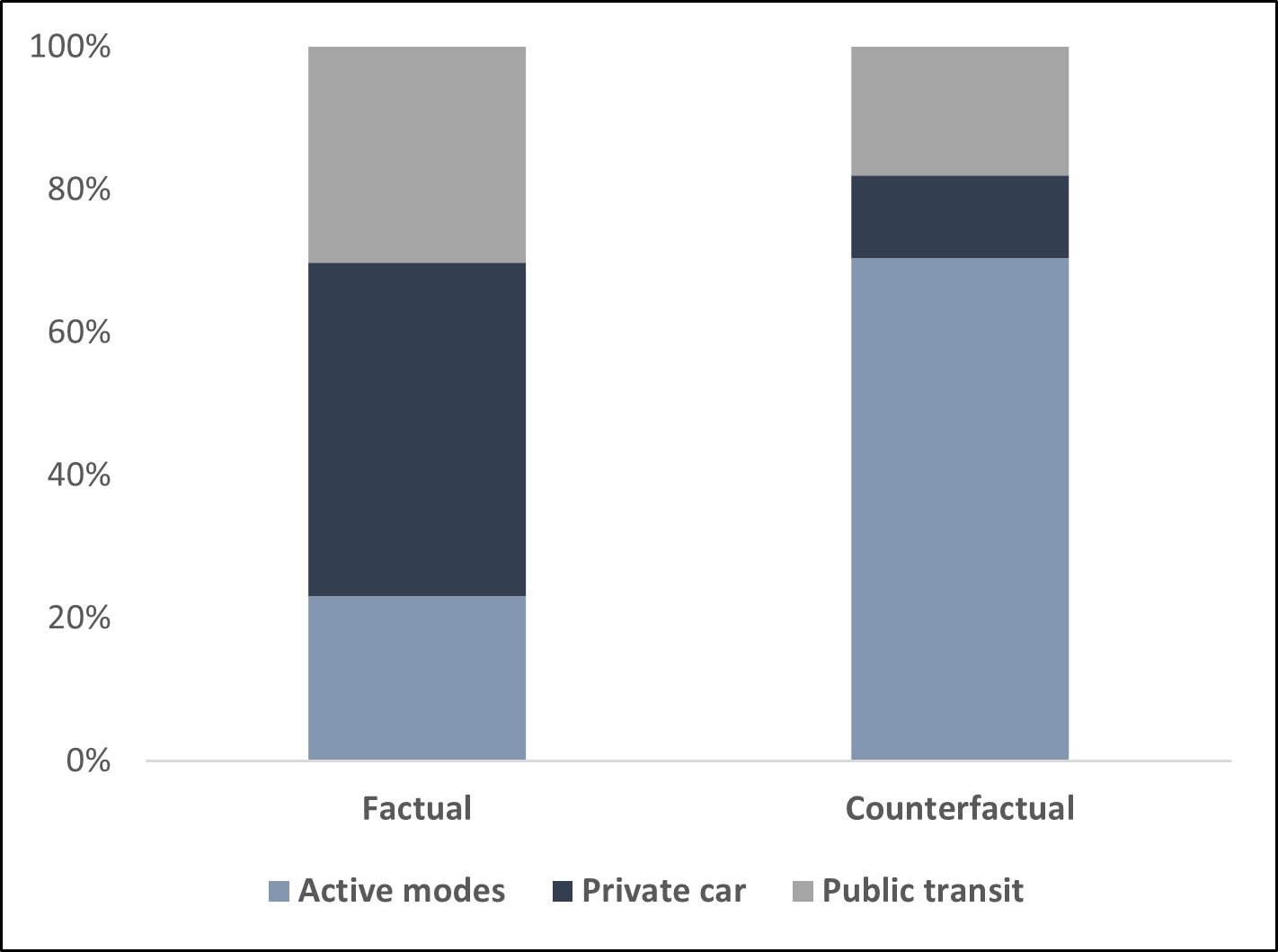}}}

    \caption{Impact of interventions on pedestrian wait time and travel mode shares}
    \label{countefactualchart}
\end{figure}

The individual treatment effect results on the validation dataset of both case studies are shown in Figure \ref{Individualcountefactualchart}. The dashed reference line represents instances where the factual and counterfactual decisions are equal. For the pedestrian crossing dataset, individuals below the reference line would experience shorter waiting times when their stress levels were controlled, while those above the line would face longer waiting times in the counterfactual scenario. It is of note that for the London travel dataset, the difference in category densities is less noticeable due to the large size of the dataset. The aggregate level results are displayed in Figure \ref{countefactualchart}, illustrating that intervening in pedestrian stress levels leads to a 35.35\% increase in the number of individuals who experience shorter waiting times on the sidewalk.

\begin{figure}[h]
    \centering
    {{\includegraphics[scale=.4]{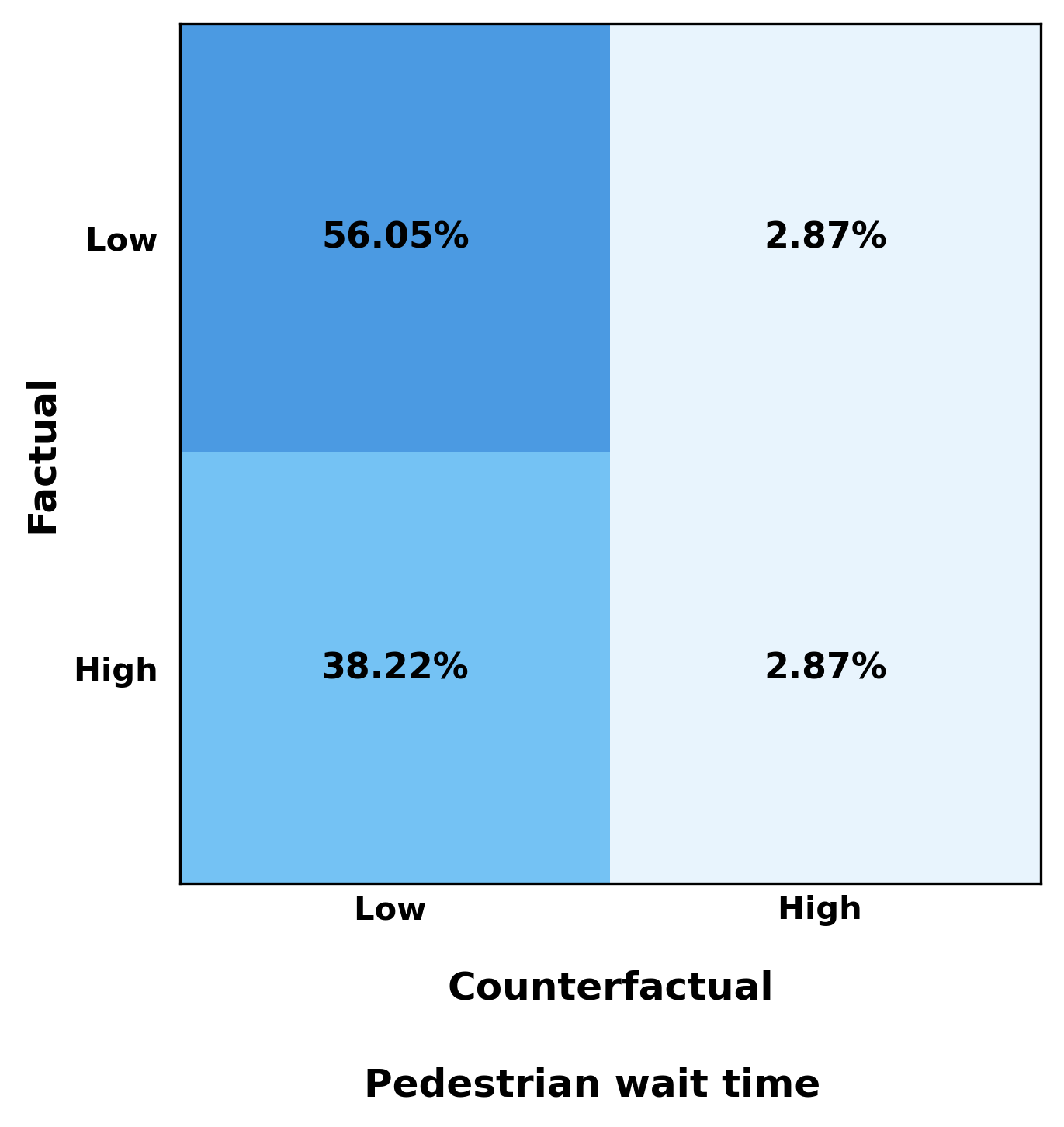}}}
    {{\includegraphics[scale=.4]{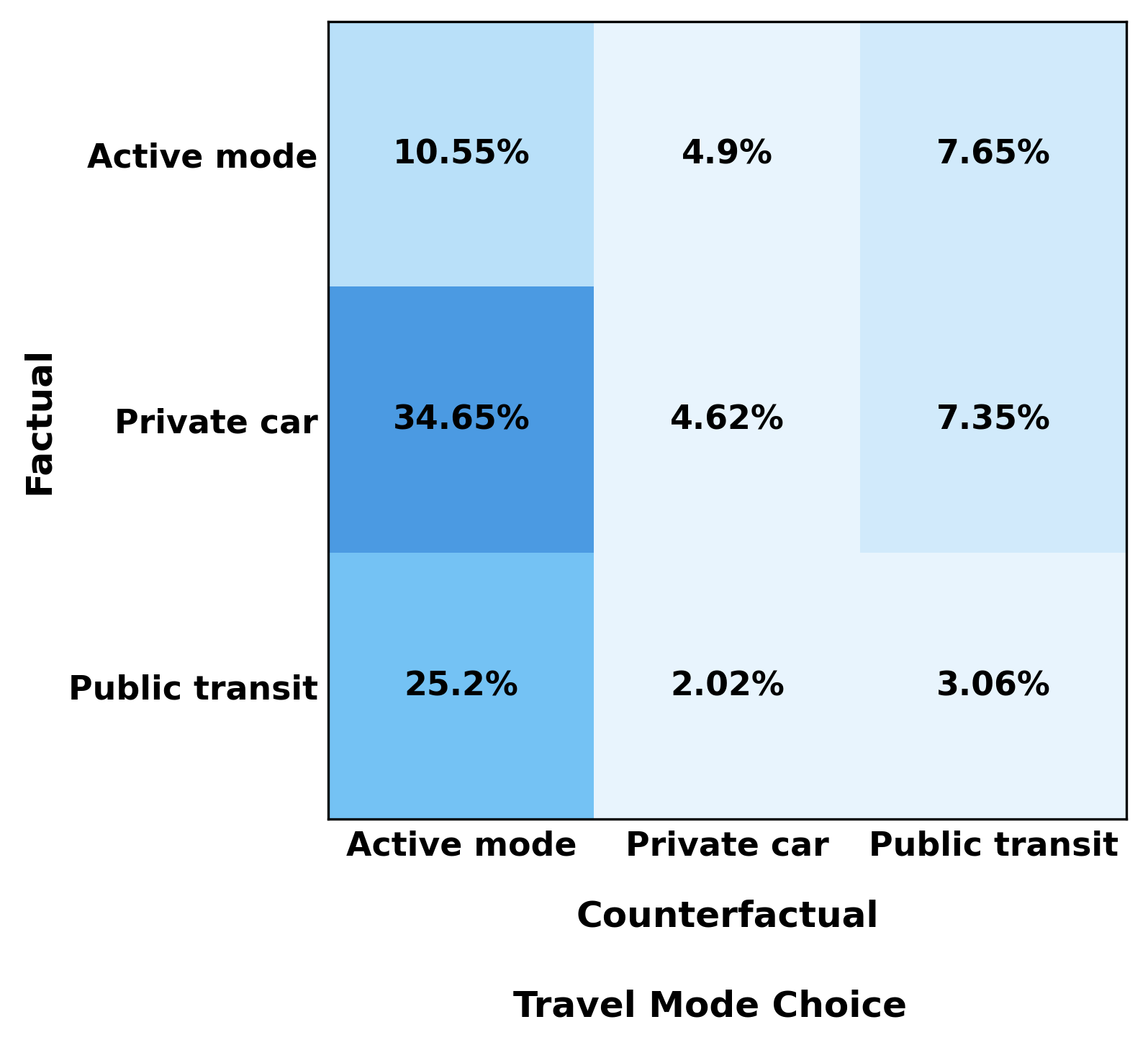}}}
    \caption{Categories distribution between factual and counterfactual world}
    \label{countefactualmatrix}
\end{figure}

 To gain deeper insights into the specific shifts, resulting from the interventions, Figure \Ref{countefactualmatrix} presents a matrix that depicts the low-high category comparison between the factual and counterfactual worlds. This matrix provides detailed information on how the intervention impacts the distribution of waiting times among individuals. The values in the matrix represent the counts of individuals who experience longer waiting times (from low to high), shorter waiting times (from low to high), and those who experience the same waiting time as in the factual world (diagonal value) in the counterfactual condition. 

\begin{figure}[!h]
  \begin{center}
   \includegraphics[width=0.5\textwidth]{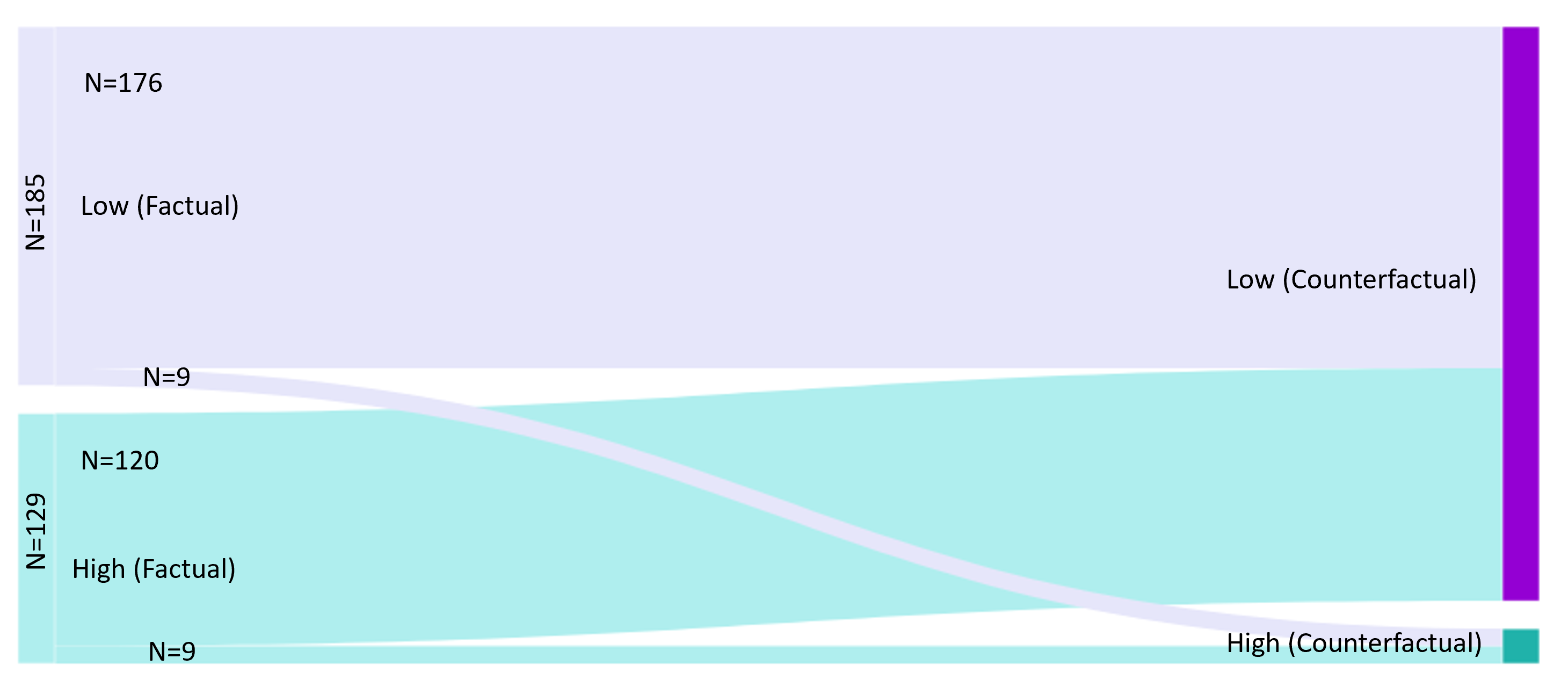}
   \caption{Wait time categories transition in factual and counterfactual worlds}\label{sankey_pedestrian}
   \end{center}
\end{figure}

\begin{figure}[!h]
  \begin{center}
   \includegraphics[width=0.5\textwidth]{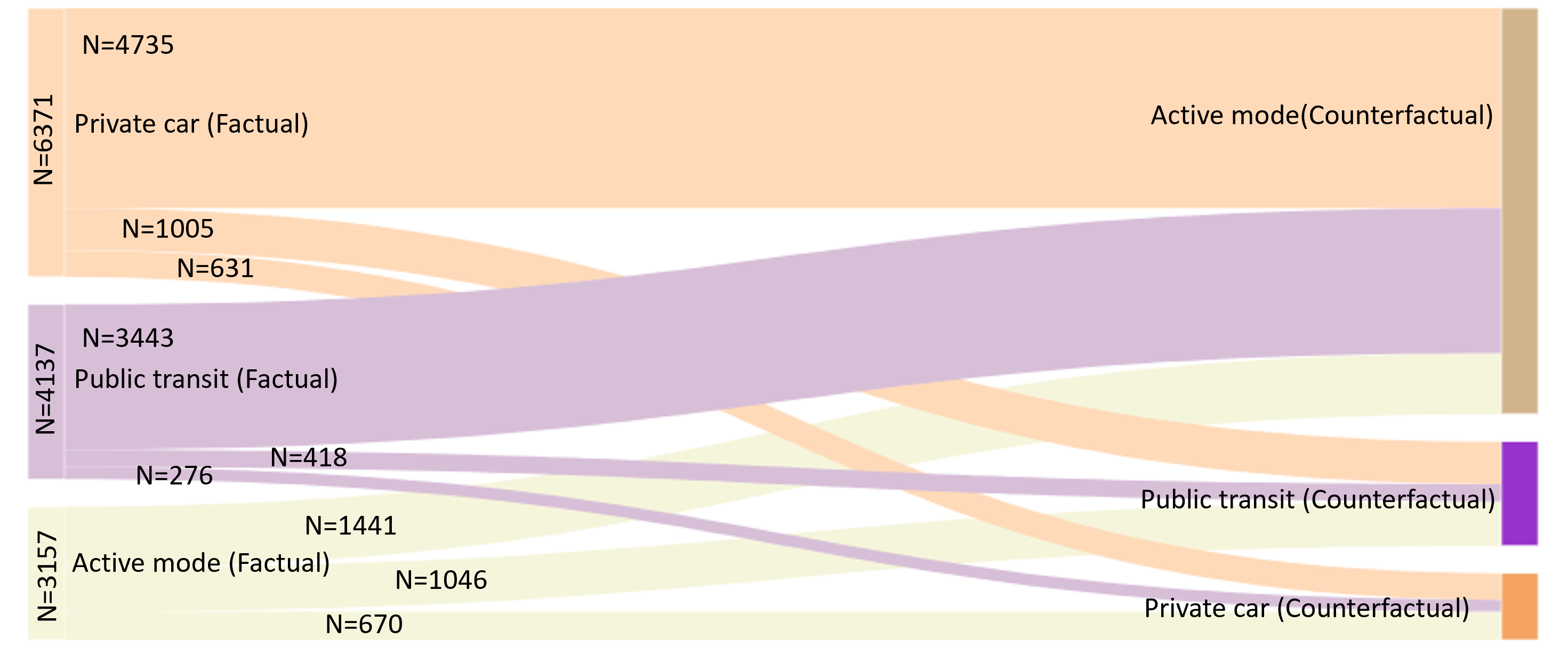}
   \caption{Travel mode choices transition in factual and counterfactual worlds}\label{sankey_london}
   \end{center}
\end{figure}

Concerning the travel distance, as shown in Figure \ref{countefactualchart}, decreasing the travel distance to the medium category in the validation dataset, approximately increases the share of active modes such as biking and walking by 47.29\%. Notably, based on the distribution of the mode share matrix (Figure \ref{countefactualmatrix}), 34.65\% of these individuals transition from using private cars to active modes. In contrast, a total of 6.92\% individuals in the dataset, in the counterfactual world would opt for private car usage, whereas in the factual world, they choose alternative travel modes. Figure \ref{sankey_pedestrian} and \ref{sankey_london} provide better insights into the transition between categories in both factual and counterfactual scenarios. For instance, Figure \ref{sankey_london} illustrates the proportion of each category within the active mode shares in the counterfactual condition.

\section{Conclusion}
\label{conclusion}
This study focuses on addressing observed and unobserved confounders in travel behaviour analysis by applying deep learning and causal inference techniques. In addition, the study presents a novel forcasting analysis for counterfactual scenarios. The study proposes the \emph{deep CAusal infeRence mOdelling for traveL behavIour aNAlysis (CAROLINA)} for policy intervention analysis. \emph{CAROLINA} includes two frameworks: Deep Structural Causal Model and Generative Counterfactual Model. The formulation of both frameworks utilizes deep learning algorithms while maintaining interpretability. We used ResNet-based models as the building blocks and \emph{CAROLINA} serves as a logical extension of ResLogit family models \parencite{wong2021reslogit, kamal2024ordinal}, while being explicitly designed for causality analysis. It is worth mentioning that our proposed model is still a probabilistic model structure that tries to discover the most probable causation with a certain degree of confidence. 

\emph{CAROLINA} learns behaviour-generating process using causal discovery methods and identifies observed confounding variables based on directed acyclic graph structures. The model also simultaneously accounts for unobserved variables through ResNet architucture. In addition, the counterfactual individual behaviour analysis facilitates the policy assessment process. This model structure assists us in making causal inferences regarding how the individual's choices would change if transportation planners intervened in specific aspects of the system. Our main purpose is to model interventional(counterfactual) distribution in such a way that a causal mechanism, related to an intervention, is changed while other mechanisms of causal structures and the latent variables remain unchanged. In the counterfactual model, we incorporate the Normalizing Flow technique into the Variational AutoEncoder framework (VAE) so as to infer the latent exogenous (noise) distribution. This component of the counterfactual model structure is specifically called Flow-based Variational Autoencoder (FVAE) in the framework. {In the context of counterfactual analysis, \emph{CAROLINA} framework uses latent abduction to guarantee that every single individual retains their true historical quirks, biases, and unobserved confounders. For policy interventions on sub-populations, the framework provides a better way to treat situations where SUTVA may not be valid.}

We evaluate the effectiveness of our method in addressing observed confounding variables using a simulated dataset. A comparison between our framework and an association-based travel behaviour model underscores the confounding impacts of collider variables. In addition, we use two distinct datasets to evaluate the performance of \emph{CAROLINA}, particularly in accounting for the impact of unobserved confounders: 1) pedestrian crossing behaviour in the presence of Automated Vehicles (AVs), collected through a virtual reality simulation, and 2) a revealed Preference (RP) London travel demand dataset. The results demonstrate that the \emph{CAROLINA} framework significantly blocks unobserved confounding associations, showing a significant improvement over the conventional structural causal model, referred to as SCM-Logit in our study.

Our analysis of the first dataset focuses on the causal relationship between stress levels and pedestrian wait times. For the second dataset, we investigate the causal mechanisms underlying individuals' choices between medium and long travel distances and how these decisions shape their overall travel behaviour. Through counterfactual analysis, we assess the impact of interventions—reducing stress levels on pedestrian wait times and altering travel distances on mode choice shares—to predict counterfactual outcomes for policy assessment. The results reveal a significant impact of intervening in pedestrian stress levels, resulting in a substantial 35.35\% increase in the number of individuals experiencing shorter waiting times on the sidewalk. Regarding the second dataset, we observe that by reducing the travel distance to the medium category, there is an approximate 47.29\% increase in the proportion of active modes, including biking and walking.

%In short, \emph{CAROLINA} framework is characterized by three fundamental properties: 

%\begin{enumerate}
%    \item Interpretability: \emph{CAROLINA} guarantees the interpretability by integrating the Residual Neural Networks or ResNets architecture into traditional behavioural models. Not only does this integration approach completely preserve the interpretability of the model while also keeping the promising predictive performance of data-driven for travel behaviour analysis.
%    \item Causal effect analysis: \emph{CAROLINA} is a Directed Acyclic Graphs (DAGs)-based model structure that is capable of identifying non-causal associations between variables and isolating the causal effect. This framework leverages the notable performance of interpretable ResNet-based model structures.
%    \item Counterfactual prediction: The modelling of the counterfactual world is another focus of \emph{CAROLINA} framework, wherein our emphasis lies in predicting human behaviour or the choice selection process after implementing specific changes. This framework is able to capture underlying latent structures that remain unchanged between the factual and counterfactual worlds.
%\end{enumerate}

While our proposed model demonstrates significant performance, it is crucial to underscore its inherent trade-offs. Firstly, in the case of certain completely unseen behaviour, the \emph{CAROLINA} framework may not operate accurately. The model is constructed based on causal structure learning from the observational dataset using causal discovery algorithms. Therefore, in unforeseen circumstances, such as the mobility behaviour during the COVID-19 pandemic, our model would suffer from a lack of reliable causal structure, labelling this causal behavioural analysis as unpredictable. 
%In addition, the measurement of unobserved confounding effects in our proposed models operates as a black box, making it difficult for modellers to derive any meaningful behavioural insights from it. Moreover, this deep learning component may also include collider unobserved confounders and conditioning on them may potentially create new confounding pathways between variables of interest. 
Another aspect of our proposed model is the setting of hyperparameters, which can considerably influence the performance of the models. Using approaches such as Bayesian search can be helpful in this regard. Furthermore, for efficient computation of the Jacobian determinant, we employ a simple flow model that necessitates a longer sequence of transformations. This choice, however, renders the model computationally more expensive. In addition, the training time of \emph{CAROLINA} framework is higher than SCM-Logit in which no deep layers are applied. For example, in the London travel dataset, the training time of our proposed model is approximately six times longer than SCM-logit. By using Graphic Processing Units (GPUs) and data reduction techniques, the computational time can be decreased, but this is out of the scope of the current work.

%Our study tries to initiate a new research area building upon the concerns developed by \cite{brathwaite2018causal}. 
In light of the findings presented in this study, several future research and potential improvements emerge. Firstly, the interaction between unobserved and observed confounders and exogenous variables can be one future direction. In addition, future studies could focus on addressing the Markov-equivalence class problem in causal structure learning and improving the accuracy of DAGs obtained through causal discovery methods. Furthermore, future researchers in the field of causality analysis can make a comparison between the performance of our proposed models and other approaches. In order to evaluate and determine the feasibility of the idea of the generative counterfactual model, we need to focus on the backcasting method in future studies, involving working backward from the future to the present by using a real temporal dataset. This approach provides valuable insights into the effectiveness of the counterfactual model structure. This evaluation helps us improve the model performance by revising the algorithms or model parameters. For example, in our counterfactual model, we utilize normalizing flow to estimate the distribution of exogenous latent variables. Yet, there may be potential for enhancing this aspect of our model by exploring other flow transformations or alternative algorithms. %Regarding the deep learning aspect of our model, it is worth noting that regularization techniques were not employed in this study to ensure a fair comparison between our proposed model and theory-driven models. However, future studies could enhance our frameworks by focusing on regularization techniques and tuning hyperparameters. 

\section*{Acknowledgements}

This research was funded by the Canada Research Chair program in Disruptive Transportation Technologies and Services (CRC-2021-00480) and NSERC Discovery (RGPIN-2020-04492) fund. We also thank Dr. Rulla Al-Haideri for providing very useful feedback, helping in coding, and serving as a sounding board during the manuscript update process.

%\newpage
%\bibliographystyle{alpha}
%\bibliography{reference}
%\bibliographystyle{plainnat}

%\bibliographystyle{apalike}
%\bibliography{reference} % Assuming your bibliography file is named "references.bib"
\section*{Appendices}
\appendix

\section{DAG: Definition, Properties, and Discovery}
\label{app:AA}
{A DAG is formally defined as follows:}

\begin{definition}[Directed Acyclic Graphs]
{A Directed Acyclic Graph (DAG), $\mathcal{G}:=(x, \zeta)$  is a graphical representation of the causal model structure, where the nodes of the graph represent random variables $(x)$, including exogenous variables, endogenous variables and unobserved variables, and the edges $\zeta$ between nodes represent causal associations or direct dependencies between variables. In this graphical model, the edges are directed in such a way that their direction represents the direction of causality and a node $x_{1}\in x$ is a parent $(Pa)$ or cause of another node $x_{2}\in x$ called child, if $x_{1}\rightarrow x_{2}$. In addition, DAGs are acyclic, meaning that there are no cycles in the graph. Therefore, each DAG consists of several causal paths or directed edges pointing away from parents or ancestors $(x_{anc})$ toward the child or descendants $(x_{des})$.}
\end{definition}
%Indeed, causal discovery enables us to identify non-causal associations between variables of interest in the process of modelling.
{To learn a DAG $\mathcal{G}$, the causal discovery consists of various techniques, including constraint-based, score-based, asymmetrical distribution, and hybrid approaches, differing in theories and assumptions} \parencite{nogueira2022methods}{. It is worth mentioning that in the literature, there is no standard causal discovery method and researchers usually select algorithms based on their studied problem and assumptions} \parencite{pearl2009causality, spirtes2010introduction, peters2017elements, nogueira2022methods}{. Sufficiency and faithfulness are the most popular assumptions in causal discovery methods, stating that:}

\begin{definition}[Sufficiency]
{There are no unobserved confounders in the causal graph $\mathcal{G}$, meaning there is an absence of unobserved common causes for all relevant variables and their shared parent variables ($Pa_i$) are fully observed in the data.}

%{There are no unobserved confounders in the causal graph $\mathcal{G}$, meaning that for all variables, all their causes $(Pa_i)$ are observed in the data.} 
\end{definition}

\begin{definition}[Faithfulness]
{Given a DAG $\mathcal{G}$, the faithfulness assumption states that the causal relationships that exist in the underlying causal graph accurately reflect the observed distribution among variables. This is denoted as:}
\begin{equation}
\label{Faithfulness}
x_1 \independent_{\mathcal{G}} x_2 \leftarrow x_1 \independent_{P} x_2
\end{equation}
\end{definition}

{Faithfulness assumption allows us to infer the causal graph, by assuming all conditional independencies in probability distribution $P$ are represented in $\mathcal{G}$.} %In the following, a brief description of four causal discovery approaches is presented.

%\begin{itemize}

%    \item 
%\subsubsection{Constraint-based approach} 
{Constraint-based approach uses conditional independence encoded in the data to infer a graphical causal structure. Constraint-based approaches are based on the faithfulness assumption and try to identify the presence or absence of certain edges in a DAG using conditional independence tests} \parencite{nogueira2022methods, spirtes2000causation}. 
%It is of note that faithfulness is a strong assumption that requires big sample sizes for conditional independence tests. 
{The PC (Peter and Clark) and Fast Causal Inference (FCI) as constraint-based causal discovery approaches are widely used in the literature. The main difference between these two algorithms is that in contrast to FCI, PC makes sufficiency assumptions.
%    \item 
%\subsubsection{Score-based approach} 
Score-based approach evaluates the goodness of fit of a given causal structure to the data and tries to optimize the goodness of fit score during learning a DAG. Score-based algorithms are also known as optimization-based algorithms. The Greedy Equivalence Search (GES) is the most popular score-based approach. This approach starts with an empty graph and iteratively adds directed edges such that the model score, like the Bayesian Information Criterion, is maximized }\parencite{chickering2002optimal}{. Fast Greedy Equivalence Search (FGES) is another modification of GES which is free from faithfulness assumption} \parencite{ramsey2017million}. 
%    \item 
%\subsubsection{Asymmetrical Distribution-based approach} 
{Asymmetrical Distribution-based approach is based on the idea that causal structure can be learned by analyzing the asymmetry in the distribution of the variables in the dataset. In other words, asymmetry is considered a fundamental characteristic of causality, and by examining these asymmetries, the method aims to uncover the underlying causal relationships. The Non-linear Additive Noise Model (NANM) and Linear Non-Gaussian Acyclic Model (LiNGAM) are classified under the asymmetrical distribution-based algorithm} \parencite{hoyer2008nonlinear, shimizu2006linear}{. Both algorithms make sufficiency assumptions.} %These methods are also known as functional causal models, where LiNGAM assumes a linear relationship with non-Gaussian noise variable while NANM is specified based on a non-linear function with additive noise. These approaches involve building models in both directions and selecting the direction that satisfies the independence criterion between the cause and the noise.
%    \item 
%\subsubsection{Hybrid approach} 
%\end{itemize}
{Hybrid approach is a combination of multiple approaches. For instance, causal Generative Neural Networks (CGNNs) as a hybrid method discover the causal relationships between multiple variables by taking into account both conditional independencies and distributional asymmetries} \parencite{goudet2018learning}.

\section{Counterfactual Modelling} \label{app:B}
\begin{figure}[!h]
  \begin{center}
   \includegraphics[width=0.5\textwidth]{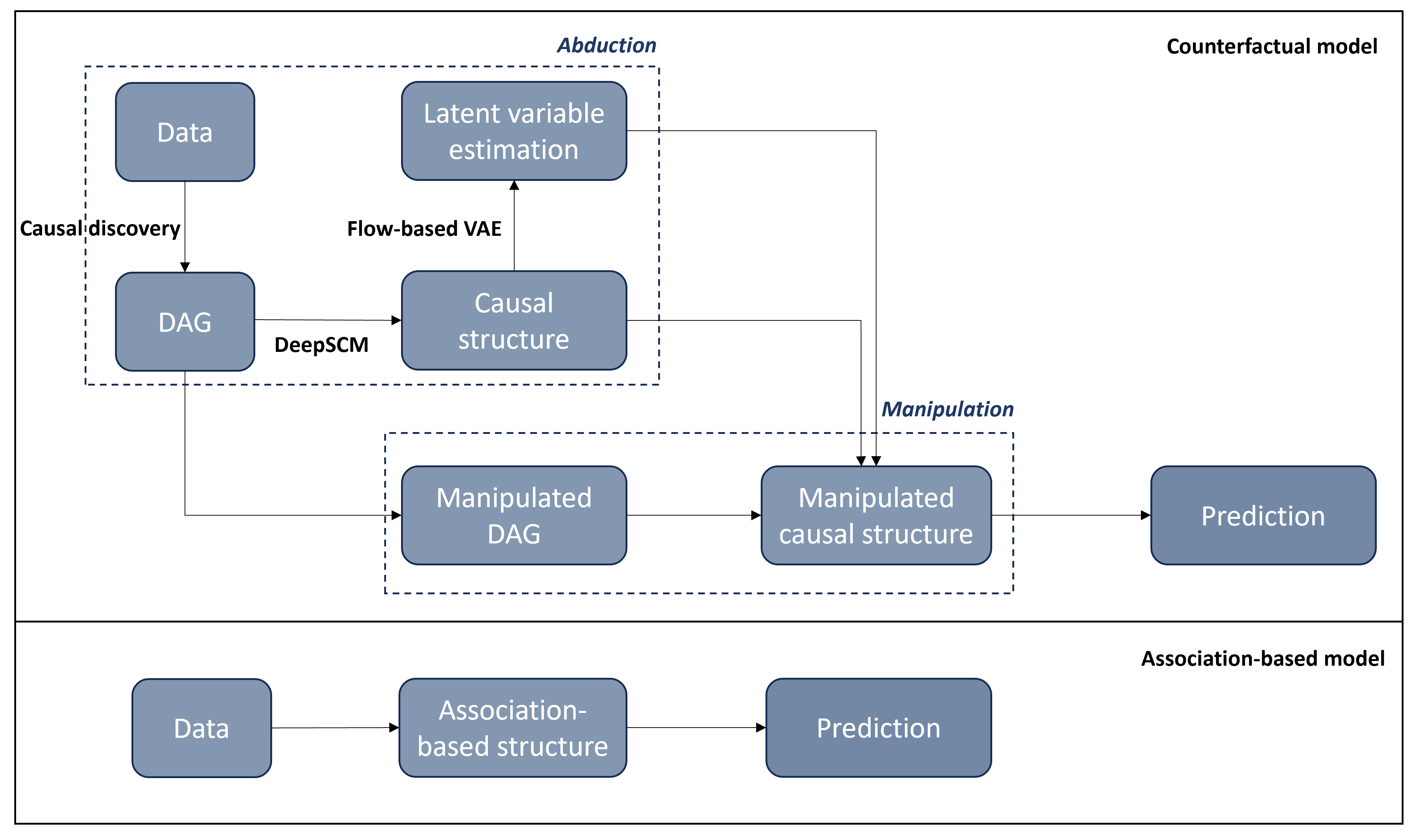}
   \caption{Counterfactual model and comparison with conventional prediction approach}\label{steps of counterfactual models}
   \end{center}
\end{figure}

Three levels of counterfactual modelling are defined as follows (also see Figure \ref{steps of counterfactual models}): 

\begin{itemize}
    \item Abduction: It refers to the process of identifying the causal graph and modelling causal mechanisms of endogenous variables. Furthermore, another aim of the abduction level is to infer the distribution of $\varepsilon_i$, which represents the likelihood of different values of the exogenous latent variables in the real world. By estimating the latent constructs, we gain insight into the values that are a partial part of generating the endogenous variables. This step helps us to understand the uncertainty and variability in the system that is assumed to be identical to the counterfactual world.
    \item Manipulation: This level involves specifying the intervention(s) or hypothetical policy and manipulating the system either in the soft or hard way, resulting in an interventional SCM corresponding to counterfactual distribution. 
    \item Prediction: The prediction level focuses on estimating the counterfactual outcome that would occur under the intervention of interest. 
\end{itemize}
%Figure \ref{steps of counterfactual models} also compares these steps with the steps involved in the association-based prediction approach.  
\section{Pseudocodes}
\label{pseudo}
The pseudocodes for the algorithms in \emph{CAROLINA} framework are outlined here. {Based on Appendix} \ref{app:B}, {Algorithm} \ref{alg:FVAE} {outlines the steps to implement the interventions (e.g., reducing distance or stress) at the individual level. In the first step of \emph{Abduction}, the Normalizing Flow is used to infer the exact independent latent traits ($\varepsilon$) of the individual based on their observed behaviour. These unobserved traits are ``frozen’’ so that the counterfactual experiment evaluates the same person, just under different conditions. In the second step of \emph{Manipulation}, the intervention (e.g., $do(\text{Travel Distance} = \text{Reduced})$) is applied on the Stage-1 DAG. It completely severs all incoming causal edges to the target variable. This isolates the variable from its natural parents, ensuring that the new value is treated as an external policy shock, rather than a natural occurrence influenced by other confounding factors in the system. In the last step of \emph{Prediction}, the targeted variable is forced to its new intervention value ($x^*$). The new downstream effects along the preserved causal pathways are computed using the estimated ResLogit structural equations, generating the final counterfactual behaviour (e.g., the choice to use a sustainable travel mode). In topological order, for each downstream node a) the new probability distribution is calculated, and b) the information is passed to the next node in the DAG until the terminal choice is reached.}

\begin{algorithm}
\tiny
\caption{Algorithm of deep structural causal model in \emph{CAROLINA}}
\label{app:A}
%\begin{adjustbox}{width=0.75\textwidth}
\begin{algorithmic}
\Require $G = (X, \xi)$ where $X$ is the set of variables and $\xi$ is the set of dependencies
\Ensure $Pa(x_i) = \emptyset$, $seq = x_{\text{outcome}}$, $x_{ex}=\emptyset$ ,$x_{des}=\emptyset$ , $x_{en}=\emptyset$
\For{$i = 1$ \textbf{to} $n$}
    \For{$j = 1$ \textbf{to} $n$, where $j \neq i$}
        \If{$I(\xi_{ji}) \neq 0$}
            \State $Pa(x_i) \gets x_j$
        \ElsIf{$Pa(x_i) = Pa(x_i)$}
        \EndIf
    \EndFor
\EndFor

\If{$Pa(x_i) = \emptyset$}
    \State $x_{\text{ex}} \gets x_i$
\EndIf

\If{$Pa(x_i) \neq \emptyset$}
    \For{$j = 1$ \textbf{to} $n$, where $j \neq i$}
        \If{$I(\xi_{ij}) = 0$}
            \State $x_{\text{des}} \gets x_i$
        \ElsIf{$x_{\text{en}} \gets x_i$}
        \EndIf
    \EndFor
\EndIf

\For{$x_{i}$ in $seq$}
    \If{$\text{visited}[x_i] = \text{False}$}
        \For{$x_{j}$ in $Pa(x_i)$}
            \If{$x_{j} \in x_{\text{en}}$}
                \State $seq[\text{len(seq)} + 1] \gets x_j$
            \ElsIf{$seq = seq$}
            \EndIf
        \EndFor
        \State $\text{visited}[x_i] = \text{True}$
    \ElsIf: Break
    \EndIf
\EndFor

\For{$x_i = \text{seq}[\text{len(seq)}]$ \textbf{to} $\text{seq}[0]$}
    \State Train ResNet-based model structures 
    %Calculate: $U^i_{kn}=:f^i_{k}(Pa^i_{kn}, \beta^i_{k}) + \varepsilon^i_{kn}$ 
    \If{$x_i$: categorical}
        \State Calculate:$P(x_i=k | Pa_{k})= \frac{\exp{(\beta_k Pa_{k} + g_{k})}} {\displaystyle\sum_{k=1}^{K} \exp{\Big(\beta_k Pa_{k} + g_{k}\Big)}}$
    \ElsIf{$x_i$: ordinal}
        \State Calculate:$P(x_i=k | Pa_{k})= \displaystyle\sigma \Bigg(\Big(\sum_{k=1}^{K} W_k (\beta_k Pa_{k} + g_{k})\Big) + b_k\Bigg)$
    \EndIf
\EndFor
\State Calculate: $\prod_{i=1}^{I} P\Bigg(x_i | Pa_i\Bigg)$
\end{algorithmic}
%\end{adjustbox}
\end{algorithm}

\begin{algorithm}
\tiny
\caption{Algorithm of counterfactual model structure in \emph{CAROLINA}}
\label{alg:FVAE}
\begin{algorithmic}
\Require $G = (x, \xi)$, $Pa(x_i)$,  $f_i(Pa_i, \beta_i)$ and $seq$
\For{each $x_i \in (seq)$} 
\State 1. Abduction: 
\State Initialize Encoder parameters $\theta_{\text{enc}}$, Decoder parameters $\theta_{\text{dec}}$, Normalizing flow parameters $\theta_{\text{flow}}$ randomly.
\For{each epoch}
    \For{each mini-batch $B$}
        \State Reset gradients.
        \For{each sample $x_n$ in $B$}
            \State Encode $x_n$ to obtain mean and log-variance parameters $\mu$, $\sigma^2$ using $\theta_{\text{enc}}$.
            \State Sample $\gamma_0 \sim \mathcal{N}(\mu, \sigma^2)$.
            \State Apply $K$ normalizing flows to $\gamma_0$, obtaining transformed $\gamma_k$ and $\log ( \text{det} |J| )$.
            \State Reconstruct $x_n'$ using $\gamma_k $ and $\theta_{\text{dec}}$.
            \State Compute reconstruction loss $(L_{\text{recon}})$ between $x_n$ and $x_n'$.
            \State Compute KL loss $(L_{\text{KL}})$ between the approximate and true latent distributions.
            \State Compute total loss $L_{\text{total}} = L_{\text{recon}} + L_{\text{KL}}$.
            \State Accumulate gradients of $L_{\text{total}}$ w.r.t. $\theta_{\text{enc}}$, $\theta_{\text{dec}}$, and $\theta_{\text{flow}}$.
        \EndFor
        \State Update parameters using the optimizer.
    \EndFor
\EndFor
\State 2. Manipulation:
\State Interventional CAROLINA: Manipulate the causal graph and accordingly CAROLINA either in the soft or hard operation
\State 3. Prediction:
\State Sample $\gamma_i \sim q_k$ 
\State Compute latent component of causal structures: $\varepsilon_i = \gamma_i - f_i(Pa_i, \beta_i)$
\State Predict $x_{counterfactual}$ for $x_{outcome}$ using the Interventional CAROLINA. 
\EndFor
\end{algorithmic}
\end{algorithm}

\section{{Economic Analysis}}\label{app:econ}
{
The Marginal Rate of Substitution (MRS) represents how much an individual $n$ is willing to trade-off between two attributes $i$ and $j$ to keep their utility for alternative $k$ without changes, providing insights into their preferences and decision-making processes related to travel. Mathematically, this definition can be written as follows:}

\begin{equation}
\label{VoTT}
\begin{split}
\displaystyle
MRS_{nij} = \frac{MU_{nki}}{MU_{nkj}}
\end{split}
\end{equation}

{where $MU$ is the marginal utility.}\\

{
We developed the marginal rate of substitution between travel distance and travel cost attributes ($MRS$) in the London passenger mode choice dataset \parencite{hillel2018recreating} based on the model in Table \ref{DSEM-London}. The $MRS$ (km/GBP) values obtained show that an individual in London is willing to pay 1GBP to avoid extra travel of 1.957km by car while 0.408km by public transit. The results show that individuals associate almost four times more disutility with the car mode compared to public transit, when it comes to longer distances. This observation aligns with the official statistics for London \parencite{tdl2025}, where the average trip length by car is 7km while for all modes it is 13km.}
%The result demonstrates that individuals who typically prefer using private cars place a higher value on their time compared to the monetary cost associated with travel. Essentially, people are willing to spend approximately four times more money to save time when using private cars. Conversely, regarding public transit, individuals are more inclined to opt for cheaper transportation modes, even if they entail longer travel times. This finding can be traced back to various factors influencing the lives of these individuals, such as financial constraints or the availability of low-cost transportation alternatives.}

\begin{figure}[h]
    \centering
    {   \includegraphics[width=0.3\textwidth]{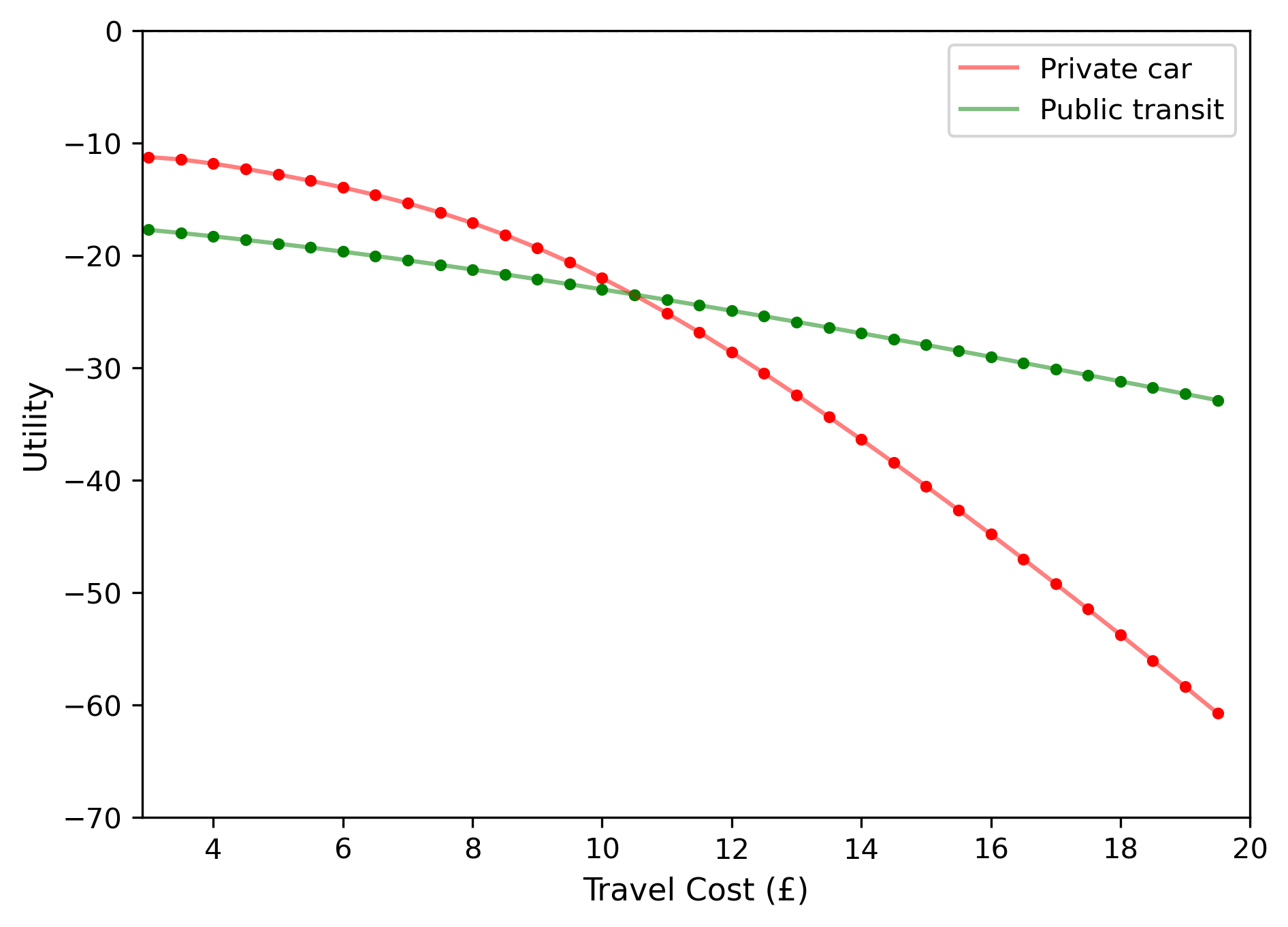}}
    {{\includegraphics[scale=.33]{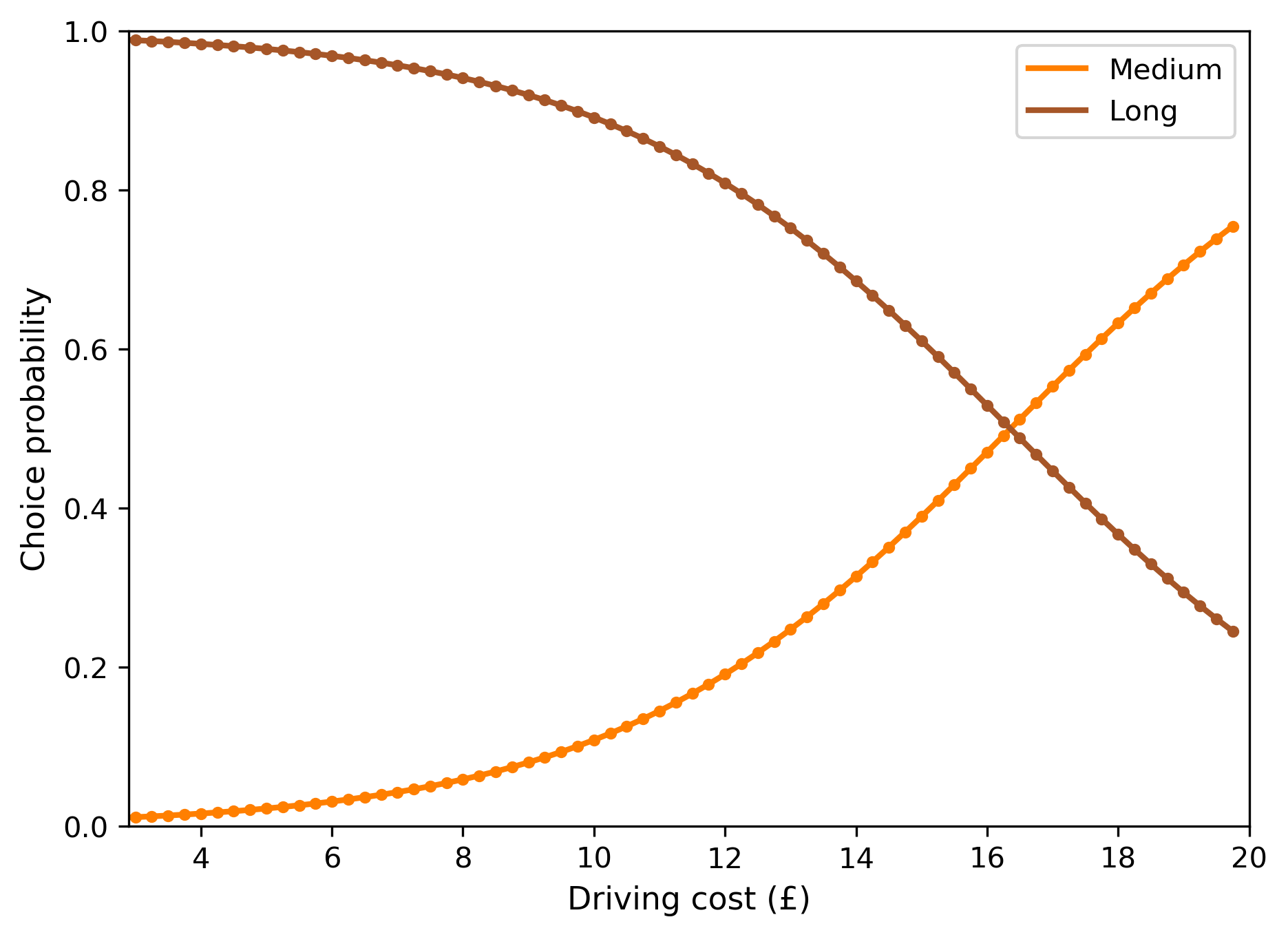}}}
    {{\includegraphics[scale=.33]{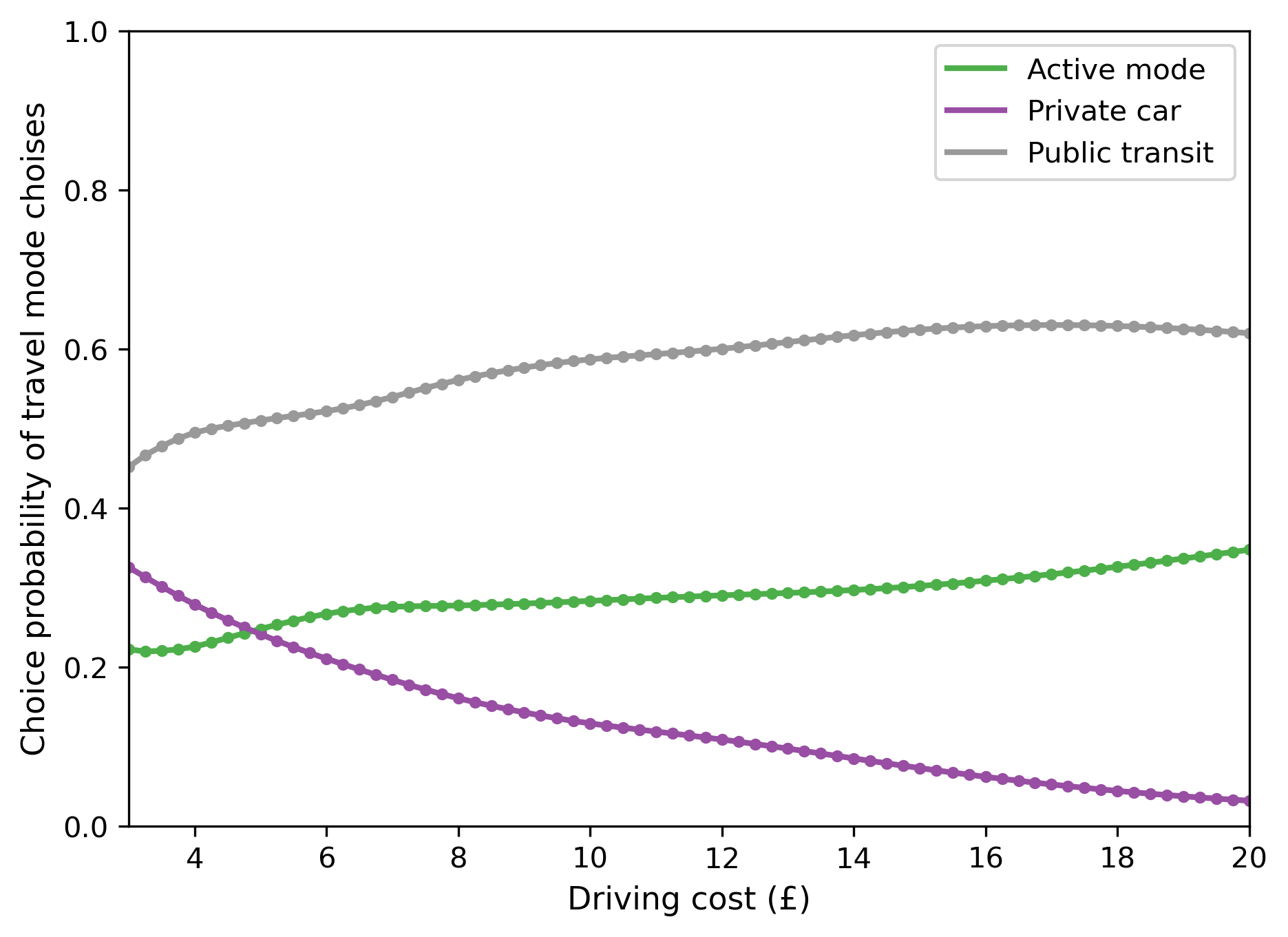}}}

    \caption{Monotonicity of marginal utility and substitution patterns}
    \label{MU}
\end{figure}

{Figure \ref{MU} shows the monotonicity of the marginal utilities and substitution patterns. As the driving cost increases, the inclination towards selecting recreational destinations gradually shifts towards more distant locations in such a way that, reaching approximately a driving cost of \pounds16, the medium-distance category emerges as a substitute for longer travel distances. Interestingly, our results also indicate that when the driving cost is less than \pounds2.5, individuals are more inclined towards opting for private car transportation in London. However, this inclination gradually changes to a preference for public transit. In fact, as the driving cost increases, there is a discernible shift in individual travel patterns in London, transitioning from long-distance travel to medium-distance travel, accompanied by a notable preference shift towards public transportation.}

\printbibliography

\end{document}